\documentclass[USenglish]{article}	
\usepackage[utf8]{inputenc}				%(only for the pdftex engine)
\usepackage[big,online]{dgruyter}	%values: small,big | online,print,work
\usepackage{lmodern} 
\usepackage{microtype}

\usepackage{amsmath,amsfonts,amssymb,amsthm,cancel,siunitx,calc,mathtools,empheq,latexsym,enumitem}
\usepackage{bm}

\usepackage{subfig,epsfig,tikz,float}	
\usepackage{graphicx}	 
\usepackage{varwidth}
\usetikzlibrary{shapes, arrows, positioning, fit, backgrounds}
\usetikzlibrary{scopes}
\pgfdeclarelayer{edgelayer}
\pgfdeclarelayer{nodelayer}
\pgfsetlayers{background,edgelayer,nodelayer,main}
\pgfkeys{/tikz/tikzit fill/.initial=0}
\pgfkeys{/tikz/tikzit draw/.initial=0}
\pgfkeys{/tikz/tikzit shape/.initial=0}
\pgfkeys{/tikz/tikzit category/.initial=0}

\usepackage{booktabs,multicol,multirow,tabularx,array}         
\usepackage{url}
\usepackage{comment}
\newcommand{\fex}{\bm{ x}}
\newcommand{\param}{ \bm {\theta} }

\newcommand{\real}{{\rm I\!R}}
\newcommand{\prediction}{ \hat{y} }
\newcommand{\classvar}{ y }
\newcommand{\classifier}{ \hat{f} }

\newcommand{\add}[1]{\textcolor{black}{#1}}

\theoremstyle{dgthm}

\theoremstyle{dgdef}

\begin{document}

%%%--------------------------------------------%%%
	\articletype{Research Article}
	\received{Month	DD, YYYY}
	\revised{Month	DD, YYYY}
  \accepted{Month	DD, YYYY}
  \journalname{De~Gruyter~Journal}
  \journalyear{YYYY}
  \journalvolume{XX}
  \journalissue{X}
  \startpage{1}
  \aop
  \DOI{10.1515/sample-YYYY-XXXX}
%%%--------------------------------------------%%%

\title{Data-driven Modeling in Metrology - A Short Introduction, Current Developments and Future Perspectives}
\runningtitle{Introduction to Data-driven Modeling in Metrology}
%\subtitle{Insert subtitle if needed}

\author*[1]{Linda-Sophie Schneider}
%\ use * to mark the author as the corresponding author
\author[2]{Patrick Krauss}
\author[3]{Nadine Schiering} 
\author[4]{Christopher Syben}
\author[4]{Richard Schielein} 
\author[1]{Andreas Maier} 
\runningauthor{L.~Schneider et al.}
\affil[1]{\protect\raggedright 
Friedrich-Alexander-University Erlangen-Nuremberg, Chair of Pattern Recognition, Erlangen, Germany, e-mail: linda-sophie.schneider@fau.de, andreas.maier@fau.de}
\affil[2]{\protect\raggedright 
University Hospital Erlangen, Neuroscience Lab, Erlangen, Germany, e-mail: Patrick.Krauss@uk-erlangen.de}
\affil[3]{\protect\raggedright 
Zentrum für Messen und Kalibrieren \& ANALYTIK GmbH, Bitterfeld-Wolfen, Germany, e-mail: Nadine.Schiering@outlook.de}
\affil[4]{\protect\raggedright 
Development Center X-ray Technology, Fraunhofer Institute for Integrated Circuits, Fürth, Germany, e-mail: christopher.syben@iis.fraunhofer.de, richard.schielein@iis.fraunhofer.de}

%\communicated{...}
%\dedication{...}
	
\abstract{Mathematical models are vital to the field of metrology, playing a key role in the derivation of measurement results and the calculation of uncertainties from measurement data, informed by an understanding of the measurement process. These models generally represent the correlation between the quantity being measured and all other pertinent quantities. Such relationships are used to construct measurement systems that can interpret measurement data to generate conclusions and predictions about the measurement system itself. Classic models are typically analytical, built on fundamental physical principles. However, the rise of digital technology, expansive sensor networks, and high-performance computing hardware have led to a growing shift towards data-driven methodologies. This trend is especially prominent when dealing with large, intricate networked sensor systems in situations where there is limited expert understanding of the frequently changing real-world contexts. Here, we demonstrate the variety of opportunities that data-driven modeling presents, and how they have been already implemented in various real-world applications.

}


\maketitle

\newpage

%Introduction
\section{Introduction} \label{sec:1}

Due to increasing digitalization, processes in industry are changing at an accelerating pace \cite{russmann2015industry, bahrin2016industry}. In the future, entirely digitized production processes are envisaged, in which complex networks of sensors, which can be adapted to specific tasks, are used for monitoring, control, and prediction \cite{kandris2020applications, landaluce2020review, majid2022applications}. In addition, there will be internet-based connectivity of the sensors, similar to the Industrial Internet of Things (IIoT) \cite{nord2019internet}. Due to the increasing availability of cost-effective IIoT-capable measurement devices, the sensor networks are much larger and more complex than in traditional measurement applications \cite{chettri2019comprehensive}. Since this also significantly increases the amount of data collected, machine learning methods are increasingly used \cite{aggarwal2021machine}. In addition, the role of modeling in the \glqq Factory of Future\grqq \ is becoming increasingly important through the use of digital twins, where a model of a physical object is updated based on an evolving data set \cite{boschert2016digital, tao2018digital, shao2019digital, liu2021review}. In light of these advancements in measurement data processing and analysis, it is crucial to re-evaluate conventional metrological modeling methodologies. Moreover, constructing accurate models of these processes is a necessity for efficiently managing production and supply chain logistics, such as ensuring quality control \add{, as it enables precise decision-making based on reliable data, minimizing errors and optimizing operational performance \cite{tercan2022machine, mahammad2023using}.} \\
Mathematical modeling is generally about translating all the relevant components of a system or process into the language of mathematics. An important part of this is the derivation of mathematical equations that specify the relationships between the components \cite{allaire2007numerical}. In the context of measurements, a distinction is made between mathematical terms that refer to known or measured quantities and those that are unknown and must be estimated from the measured data. The former includes quantities representing the response and calibration of individual sensors, as well as applied corrections to account for environmental effects. Unknown quantities comprise the stimuli to the sensors and representations of system properties derived from these stimuli \cite{ghamisi2019multisource, fayyad2020deep, yeong2021sensor, li2022deep}. \\
A model can be assembled by utilizing the governing physical laws, \add{often incorporating modular modeling approaches like port-Hamiltonian systems to ensure a comprehensive and structured representation of the underlying processes \cite{van2014port}.} Expert knowledge and the underlying theory of the process can guide the definition of dependencies among the considered quantities and identify those relevant for modeling \cite{rajani2001comprehensive, xiang2010physics, cubillo2016review}. Conversely, one can employ a data-driven model that discerns the relationships among variables based solely on data, with no underlying theoretical backing \cite{solomatine2008data, arridge2019solving}. Numerous models blend both these physical and data-driven elements, often referred to as hybrid models \cite{willard2020integrating, pulpeiro2022integration, chen2023integration}. \\
Mathematical modeling is a crucial instrument that aids metrology and applications dependent on measurements \cite{placko2013metrology}. It underpins the creation of sensors and measurement systems, allows for deductions and forecasts about relevant quantities, and enhances comprehension of actual processes or systems. A model furnishes a simplified yet abstract representation. Furthermore, models are useful to gain an understanding of the functioning and structure of the system under consideration. Such an understanding can be used to generalize the functioning of the system to other scenarios \cite{rummukainen2010state, keeling2005networks, ellis2014tutorial}. Thus, models can form the basis for the simulation of a system. In systems theory, for example, models are the basis for the description, analysis, and synthesis of dynamic system behavior \cite{ashby1957introduction, bertalanffy1968general, wiener2019cybernetics}. Crucially, models provide the basis for the evaluation of uncertainty. \add{Faulwasser et al. \cite{faulwasser2023behavioral} provide a comprehensive review aimed at enhancing systems theory by integrating stochastic elements into the behavioral approach, drawing upon classical concepts from Willems and Wiener \cite{willems2007behavioral, 80d13487-f58d-3456-b56d-85e1178fd369}. Their review aims to establish a robust framework for data-driven control and analysis of stochastic systems, bridging theoretical underpinnings with practical applications.}  This is necessary to understand the quality of estimates of quantities, to use these estimates for decision making and to ensure the traceability of measurement results \cite{laner2014systematic}. \\
The current trend towards the digital transformation of the manufacturing industry is driven by the use of large-scale networks, so-called \glqq smart \grqq \ sensors and artificial intelligence to automatically make decisions about and control production processes \cite{landaluce2020review, majid2022applications}. This change brings several challenges for modeling. In particular, the lack of physical models is compensated by the availability of large amounts of sensor data, leading to a dependence on data-driven models \cite{solomatine2008data, arridge2019solving}. However, it is still necessary that the decisions of a data-driven approach are explainable and understandable \cite{arrieta2020explainable}, and that the evaluation of uncertainty is possible in order to ensure trust in the measurement results \cite{klas2018towards}. Further challenges arise from sensor redundancies \cite{gao2003analysis, curiac2009redundancy, nguyen2021mobile}, synchronisation problems \cite{ganeriwal2003timing, sivrikaya2004time, maggs2012consensus} or sensors of different quality \cite{ghosh2008coverage, fan2010coverage, meguerdichian2001coverage}, among others \cite{roumeliotis2002distributed}. \\
Here, we provide an overview of the application of traditional analytical parametric modeling to dynamic and distributed measurement systems. Subsequently, the application of data-driven modeling to such systems is motivated, and the different methods of data-driven modeling are discussed. Finally, intelligent, adaptive systems are discussed and the concept of digital twins is described in detail.

% Modelling in Metrology
\section{Modeling in Metrology} \label{sec:2} 

\subsection{Aims of Modeling} \label{sec:2.1}

A model is understood to be a representation of reality reduced to its essential components - usually the representation of a process or system. A model can be used to determine a measurement result and to explain, simulate or evaluate a process or system. In addition, a model can also be used to design a new sensor and, with its help, systematically and comprehensibly design and build a system or process flow \cite{xiang2010physics, arridge2019solving, solomatine2008data}.

To understand and quantitatively assess complex systems and processes, requires to simplify and abstract them significantly \cite{boccara2010modeling, ottino2004engineering, newman2011complex, liu2016control, thurner2018introduction}. This leads to the inherent limitations and imperfections of such models, constraining their ability to depict real-world behaviors accurately. However, it's critical to ensure these models consider all vital elements and factors impacting the intended results. In general, a model should meet three major criteria \cite{schilling2023predictive}: First, a model should be falsifiable, i.e. there should exist experimental paradigms to assess a candidate model. This is in line with Popper’s ideas \cite{popper1963science}. Second, a model should make quantitative predictions, as opposed to purely qualitative, predictions \cite{lazebnik2004can}. Third, a model should be as simple as possible, i.e., contain the smallest possible number of parameters and assumptions. Hence, if two models explain a given system or process equally well, the simpler one is considered to be the better one, a principle referred to as Ockham’s razor \cite{lazar2010ockham}. 

Generally, a model consists of a set of mathematical equations involving at least two quantities of interest. Furthermore, a model can generally be represented graphically, in tabular form, as a flow chart or schedule, as explanatory text, or in other ways \cite{schilling2023predictive}.

\subsection{Modeling of Complex Systems} \label{sec:2.2}

Modeling of complex systems requires a process of system reduction, abstraction, and decomposition that, in particular, involves the following steps \cite{kozin1986system, ljung2010perspectives, keesman2011system}:

First, \emph{delimitation} involves defining the scope of applicability for the model. Models usually apply to specific situations or systems, and the conditions under which they apply must be clearly stated.

Second, \emph{abstraction} which means that the model is developed to apply not only to a very specific, individual or even unique system but instead to a class of systems that behave in a similar way. Typically, this process involves parameter adjustments to ensure the model can cover a range of similar systems.

Third, in the \emph{reduction} step, the model is simplified to make it more understandable. This comprises eliminating details or variables that have little influence on the system's behavior. Usually, this process leads to deviations from the real-world behavior of the system, which must be evaluated and accounted for.

Fourth, in the step of \emph{decomposition}, the complex systems are broken down into simpler subsystems, which can be more easily modeled. The challenge here is to define the boundaries of these subsystems so that they form meaningful functional units that are easier to understand and model \cite{lazebnik2004can}.

Subsequently, in a process called \emph{aggregation}, the sub-systems (or sub-models) created in the decomposition step need to be combined and re-assembled again in order to form a complete model of the system under consideration. This process must ensure that the sub-models fit together via adequate interfaces to form a valid overall-model. Any deviations or errors occurring at this stage may require revising the model.

Finally, in the \emph{verification} step, the resulting model is empirically tested against real-world data, with the aim of identifying and evaluating any discrepancies between model and reality. This may involve adjusting the model to match the desired capabilities, taking into account system dynamics, time variability, and response to external disturbances.

The described modeling approach is particularly suitable for complex systems where understanding the interactions between various sub-systems is crucial. It also highlights the iterative nature of modeling, where the model is continually tested, revised, and refined to improve its accuracy and reliability \cite{schilling2023predictive}.\\
\add{
In metrology, modeling complex systems involves simplifying complicated phenomena in order to better understand, predict, and control their behavior. To illustrate the steps in this process, we examine a study on mathematical and physical modeling of complex biological systems \cite{Sacco2019Mathematical}. The model focuses on the simulation of biological fluid flow in ophthalmology, with specific conditions outlined for its applicability. Abstraction transforms complex fluid dynamics into mathematical equations, facilitating the study of flow patterns in similar systems. Reduction simplifies structures and interactions into key variables for clearer understanding. Decomposition divides the system into manageable subsystems, while aggregation integrates these models to capture overall system behavior. Verification involves testing the integrated model against experimental data to improve predictive accuracy.
}

\subsection{Measurement Methods and Model Structure} \label{sec:2.3}

\subsubsection{Forward Models and Inverse Models}

The traditional (cognitive-systematic) method of understanding a system or process follows a path from cause to effect, or, in terms of time, from the beginning to the conclusion of a process \cite{sommer2005systematic}. This method is frequently used to create explanations, justifications, educational content, and more. It mirrors the prevalent modeling approach in both science and engineering. Consequently, it is the standard mode of depicting measurement systems and the progression of signals, starting from the cause (the metric with its mostly undetermined value) to its outcome, which is the shown or resultant signal value of the measuring system \cite{bertalanffy1968general}.

In measurement technology, the initial step often involves a sequence of measurements. The individual components, derived from breaking down the complete measurement, are known as transmission elements. The quantity to be measured is the input, the first in the series of transmission elements are sensors or transducers, and the sequence concludes with the display or signal output. Interfering or disruptive factors might be viewed as secondary input variables. This cause-and-effect model of a measurement is useful for simulating the core and critical behavior of a developing sensor or measurement system \cite{bentley2005principles, doebelin2007measurement}.

%At this point, the cause-effect modeling approach is to be illustrated by the example of the dimensional measurement process. The individual transmission elements can initially be considered separately from each other. This consideration incorporates both known physical correlations for the transmission behaviour, e.g. to take into account optical surface properties of the test object, and empirical correlations, e.g. the thermal expansion or the temperature dependence of the measurement object. Such a cause-effect model of a measurement can be used, for example, to simulate the essential and significant behaviour of a sensor or a measurement system under development.

In the realm of metrology, forward modeling, also known as cause-effect modeling, typically relies on foundational physical principles and associated material measurements to produce or replicate reference values for specific quantities. While the primary objective in metrology is to ascertain the most precise value of a quantity and gauge its uncertainty, this often involves tracing back from the displayed output to the original measured value. A reverse model facilitates the extraction of details about the initial measurement quantity from the presented value, insights into potential major influencing factors, and the modeled procedure. Thus, measuring assessment entails addressing what is termed an inverse problem \cite{beck1998inverse, groetsch1993inverse}. The resulting inverse model is also called measurement equation \cite{vim}, evaluation model or measurement model \cite{gum}. It is obtained by inverting the cause-effect model or the forward model, respectively, and is the starting point for applying the uncertainty assessment methods described in the Guide to the Expression of Uncertainty in Measurement (GUM) \cite{gum} and its supporting documents \cite{gum1}. Data-driven models, in particular adaptive and learning models, can usually be taught directly in the "inverse direction". Whereas inversion of forward models is generally possible, it often may result in numerically unstable equation systems. To reduce this problem, regularisation techniques allow for more robust inversion procedures \cite{vogel_2007}.

\subsubsection{White-, Grey- and Black-Box Models}

Beyond categorizing models based on their directionality into forward and inverse, another differentiation can be drawn considering the transparency-opacity spectrum, leading to white-box, grey-box, and black-box models \cite{khan2012comparative, yang2017investigating, pintelas2020grey, lo2020identification, loyola2019black}. 

\begin{figure}[tb]
	{ \hspace*{-4mm} \footnotesize \begin{tikzpicture}
			% Define block styles
			\tikzstyle{block} = [rectangle, node distance=1.75cm, minimum width=0.994\textwidth, minimum height=0.5cm, draw, fill=white, 
			text width=20em, text centered, minimum height=1.5em, inner sep=0pt]
			\tikzstyle{block2} = [rectangle, node distance=1.75cm, minimum width=0.994\textwidth, minimum height=0.5cm, draw, fill=white, 
			text width=20em, text centered, minimum height=2cm, inner sep=0pt]
			\tikzstyle{blockg} = [rectangle, minimum width=0.25\textwidth, draw,  
			text width=11.5em, rounded corners, minimum height=6cm, inner sep=2pt]
			\tikzstyle{line} = [draw, -latex']
			\tikzstyle{invis} = [draw, fill=yellow!10, node distance=4.25cm, 
			minimum height=2em]
			\tikzstyle{invisg} = [draw, fill=yellow!10, node distance=3.5cm, 
			minimum height=2em]
			\tikzstyle{matheq} = [node distance=8.75cm, text width=21em, minimum width=1cm, 
			minimum height=2em, text centered]

			% Place nodes
			\node [block] (second) at (-1,0) {Modeling in Metrology};
			\node [blockg, fill=white, ,text depth = 3.9 cm] (third) at (-5.48,-4.35) {
				{\begin{varwidth}{\linewidth} 
						\underline{Input:} \begin{itemize}[leftmargin=*]
							\item Data with unknown meaning
						\end{itemize}
						\underline{Model:}
						\begin{itemize}[leftmargin=*]
							\item Known physical relations
							\item Prior knowledge
			\end{itemize}\end{varwidth}}};
			\node[above= 0.1cm of third,align = center, text width=15em] (white) {Analytical Modeling\\ (White-Box Modeling)};
			\node [blockg, fill=black, right = 0.6cm of third, ,text depth = 3. cm] (fourth) { \color{white}
				{\begin{varwidth}{\linewidth} \underline{Input:} \begin{itemize}[leftmargin=*]
							\item Data with limited known meaning
						\end{itemize}
						\underline{Model:}
						\begin{itemize}[leftmargin=*]
							\item Arbitrary functional relations
							\begin{itemize}[nosep]
								\item mathematical
								\item statistical
								\item learned
								\item neuronal
							\end{itemize}
			\end{itemize}\end{varwidth}}};
			\node[above= 0.1cm of fourth, align = center, text width=15em] (black) {Data-driven Modeling \\ (Black-Box Modeling)};
			\node [blockg, fill=gray, right = 0.6cm of fourth, ,text depth = 2.7 cm] (fifth){ \color{white}
				{\begin{varwidth}{\linewidth} \underline{Input:} \begin{itemize}[leftmargin=*]
							\item Data with unknown or only partially known meaning
						\end{itemize}
						\underline{Model:}
						\begin{itemize}[leftmargin=*]
							\item Mixture of physical and functional relationship
							\begin{itemize}[nosep]
								\item mathematical
								\item statistical
								\item learned
								\item neuronal
							\end{itemize}
			\end{itemize}\end{varwidth}}};   
			\node[above= 0.1cm of fifth,align = center, text width=15em] (grey) {Mixed Modeling \\ (Grey-box Modeling)};
			
			\node [block2, below= 0.6cm of fourth] (end)  {
				{\begin{varwidth}{\linewidth} \underline{System characterisation} 
                        \begin{itemize}[leftmargin=*, nosep]
							\item Measurement uncertainties
							\item Reproducibility and comparability
							\item Bayesian description of rules
							\item Explainability of decision making
						\end{itemize}
			\end{varwidth}}};
			
			% Draw boxes
			\begin{scope}[on background layer]
				\node[draw, fill=white, fit= (white) (third), inner sep=1pt,inner xsep=-3.5mm] (firstBox){}; %, inner ysep=5mm
			\end{scope}
			\begin{scope}[on background layer]
				\node[draw, fill=white, fit= (black) (fourth), inner sep=1pt,inner xsep=-3.5mm](secondBox){}; %, inner ysep=5mm
			\end{scope}
			\begin{scope}[on background layer]
				\node[draw, fill=white, fit= (grey) (fifth), inner sep=1pt, , inner xsep=-3.5mm](thirdBox){}; %, inner ysep=5mm
			\end{scope}

			\coordinate[below = 0.01cm of firstBox] (a1); 
			\coordinate[below = 0.01cm of secondBox] (b1);  
			\coordinate[below = 0.01cm of thirdBox] (e1);
			
			\coordinate[below = 0.58cm of firstBox] (a2); 
			\coordinate[below = 0.58cm of secondBox] (b2);  
			\coordinate[below = 0.58cm of thirdBox] (e2);
			
			% Draw edges
			\path [line] (a1) -- (a2);
			\path [line] (b1) -- (b2);
			\path [line] (e1) -- (e2);
	\end{tikzpicture}}
	\caption{White, grey and black box models with their basic properties \cite{paper}.}
	\label{fig:tmp}
\end{figure}
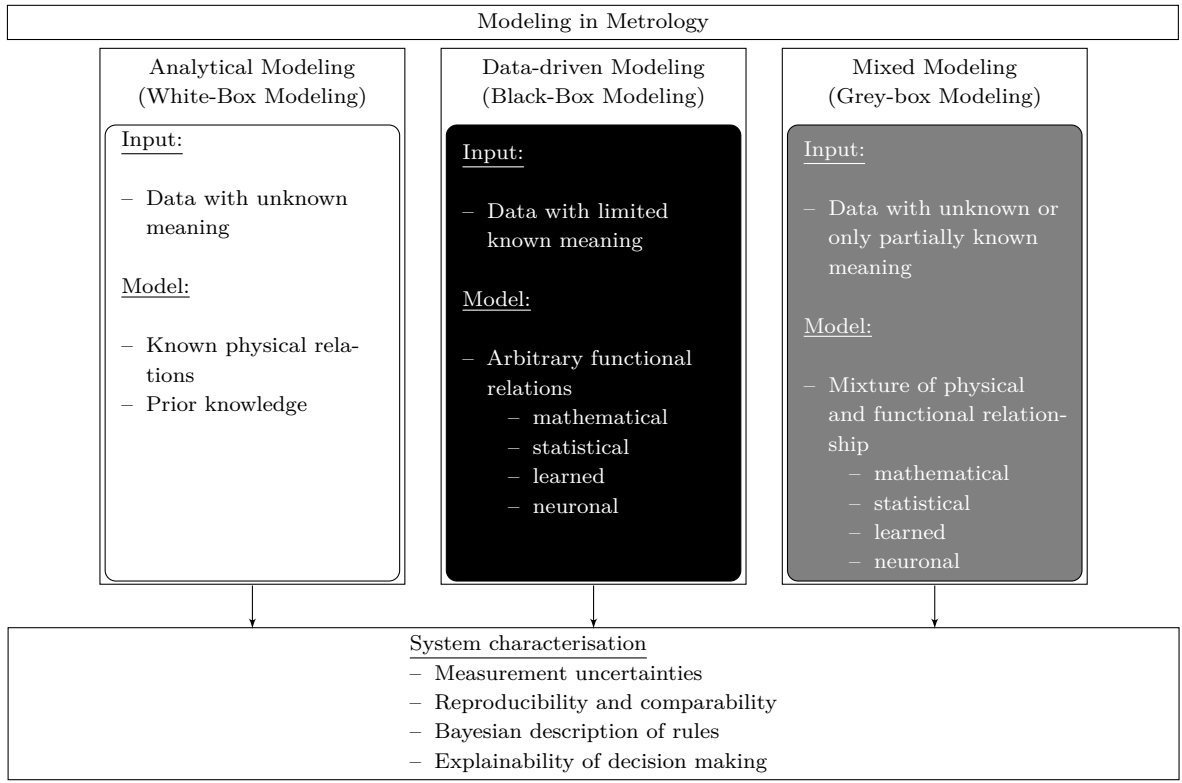

The mathematical terms used to represent the relevant components or quantities of a system or process generally include model parameters that serve to describe the dependencies between these terms. The nature and significance of such parameters depend on the nature of the model. For example, in a model derived from physics, the model parameters usually refer to specific physical properties, such as temperature coefficient, mass or time. Such models are called white-box models \cite{loyola2019black}. Sometimes, however, the dependencies cannot be derived from physical arguments alone. For example, linear interpolation through a point cloud may be derived from correlation arguments rather than purely physical considerations. However, the parameters obtained in this way can still carry a physical interpretation. Models of this type are thus a mixture of a direct relationship to a known quantity (physical or otherwise) and a more abstract relationship. Such models are called grey-box models \cite{rogers2017grey, leifsson2008grey, worden2007identification, pitchforth2021grey, hellsen2000grey}. If there is no direct relationship derived from physical or other arguments, the model is usually called a black-box model \cite{worden2007identification, loyola2019black}. In Figure \ref{fig:tmp} an overview of these three types of models is provided. 
The general principles described above apply equally to white-, grey- and black-box models. Depending on the choice of method, generalization and transferability can be challenging, as generally, learning only guarantees stability if the training data is drawn from the same distribution as the test data. In particular, parameter-rich black-box models typically suffer from such problems, as described in section \ref{sec:4}. To compensate for such data maldistribution, the discipline of transfer learning or adaptation has been developed, which can contribute to the generalization of black-box models \cite{pan2009survey, torrey2010transfer, weiss2016survey, day2017survey, tan2018, zhuang2020comprehensive, iman2023review}. However, domain transfer is more difficult than classical modeling methods \cite{ganin2015unsupervised} due to the limited interpretability of the learned parameters \cite{pan2009survey}.

% Analytical Modelling

\section{Analytical Modeling: White-Box Models} \label{sec:3}

\subsection{Overview} \label{sec:3.1}

Analytical modeling is essentially the study of the structure of a system and is therefore also known as structural modeling \cite{kozin1986system}. The aim is to represent the (physical) properties of the system, including all relevant influencing factors and interactions, as accurately as possible in a mathematical model \cite{ljung2010perspectives}. The result is a model with fully known properties, often referred to as a white box model. Typically, the quantities used for such modeling are tangible physical quantities such as pressure, temperature, flow rate, irradiance, or angle \cite{keesman2011system}. \add{
Moreover, physics-based approaches like port-Hamiltonian systems offer a unified framework for modeling complex systems across various physical domains in analytical modeling. These approaches emphasize energy as a universal language, enabling robust analysis and control across diverse domains such as mechanical, electrical, and hydraulic systems \cite{van2014port}.}

The construction of an analytical model requires an adequate physical understanding of the system and all its influencing elements. However, unknowable or unobservable influences can be represented as random variables, serving as a form of model reduction \cite{gardner1983parameter, klir1995fuzzy, mula2006models, aien2016comprehensive, jiang2018probability, acar2021modeling}. This modeling approach inherently involves aspects of decomposition and aggregation through the chosen selection and merging of the subsystems used \cite{aastrom1971system}.

This approach is useful when the components of a system, together with their properties, can be explicitly defined for the synthesis of a new system. The basic behavior of the system is then directly determined by the selected system components, their respective behavior, and the way they are combined (e.g., by a block diagram). Finite element models, where the interaction between elements is represented by specific physical relationships, also fall into the category of analytical models \cite{hughes2012finite}.

While the fundamental behavior (e.g. dynamics) of the system is determined by specifying the system structure, an analytical model can be tuned to physical reality by modifying model parameters (e.g., gains, time constants). The identification of these parameters can be achieved through specific experiments (such as the application of impulse or step functions or harmonics) or during system operation by comparing expected and actual system responses. If this step of parameter adjustment fails to produce the desired quality of system modeling, structural modifications can be applied to the model to account for the causes of deviations that have not been previously considered \cite{astrom1994adaptive, box2015time, nise2020control}.

Prior knowledge of the system structure is essential for the development of analytical models. The level of detail in the known system structure determines the accuracy achievable by the model. In addition, incorporating prior knowledge of the model parameters can be beneficial; typically, empirical optimization of these parameters (in a data-driven manner) improves the quality of the model fitting \cite{astrom1994adaptive, box2015time, nise2020control}.

\add{Projection-based model reduction methods complement the principles of white box modeling by simplifying detailed, physics-based models into more computationally efficient forms while maintaining their interpretability and accuracy \cite{benner2015survey}. These reduction methods start with high-fidelity models grounded in the system's physical laws and use projection techniques to create lower-dimensional models that still reflect the essential dynamics and parameters of the original systems. This process preserves the transparency and interpretability inherent in white box models, allowing for meaningful insights into the reduced model's behavior in relation to known physical principles. Moreover, just like in white box modeling, these reduced models undergo rigorous verification against full-scale models or empirical data to ensure their fidelity, making projection-based model reduction a natural extension of white box modeling that balances detail with computational efficiency.}

\subsection{Uncertainty Evaluation}
\add{In white box modeling, uncertainty evaluation typically involves deterministic methods where uncertainties in input parameters are propagated through the model to assess their impact on the output. This process often utilizes sensitivity analysis, which quantitatively determines how changes in input variables influence the model's predictions. A prominent method in this category is the application of Taylor series expansions to approximate the effect of input uncertainties on the output, allowing for a clear understanding of how each input contributes to the overall uncertainty \cite{Wang2005Propagation}. This approach provides a direct link between the model inputs and outputs, facilitating the identification and minimization of key sources of uncertainty. An example of this method in metrology is the development of a simulation framework designed to evaluate measurement uncertainty in detailed real-life measurements, supporting structured modeling scenarios and introducing various techniques to optimize these scenarios, as discussed by Wolf \cite{Wolf2009A}.}

\subsection{Advantages and Limitations} \label{sec:3.2}

The advantages of analytical modeling are primarily in its interpretability, as the quantities and functional blocks involved typically allow a tangible physical interpretation \cite{klir1995fuzzy}. This often allows an easy plausibility check of the fundamental model properties. By representing specific physical system properties, analytical models often facilitate an accurate system description over a wide range of values of the modeled quantities \cite{astrom1994adaptive, banks2010discrete}. Of course, the validity range defined in the delimitation step of model generation must be taken into account \cite{saltelli2008global}.

A disadvantage of analytical models is the requirement for a comprehensive physical understanding of the system in terms of its structure, function and any subsystems present, which typically requires specialized knowledge of the system \cite{simpkins2012system}. If the structure of the system isn't known in advance, considerable effort may be required to analyze the system and parameterize the model \cite{box1987empirical}. This is essential in order to identify relevant influencing factors and cause-effect relationships or to carry out appropriate experiments \cite{astrom1994adaptive, franklin2002feedback}.

% Data-driven Modelling

\section{Data-driven Modeling: Black-Box Models}\label{sec:4}

\subsection{Overview}\label{sec:4.1}

\begin{figure}
	\centering
	\includegraphics[width=\linewidth]{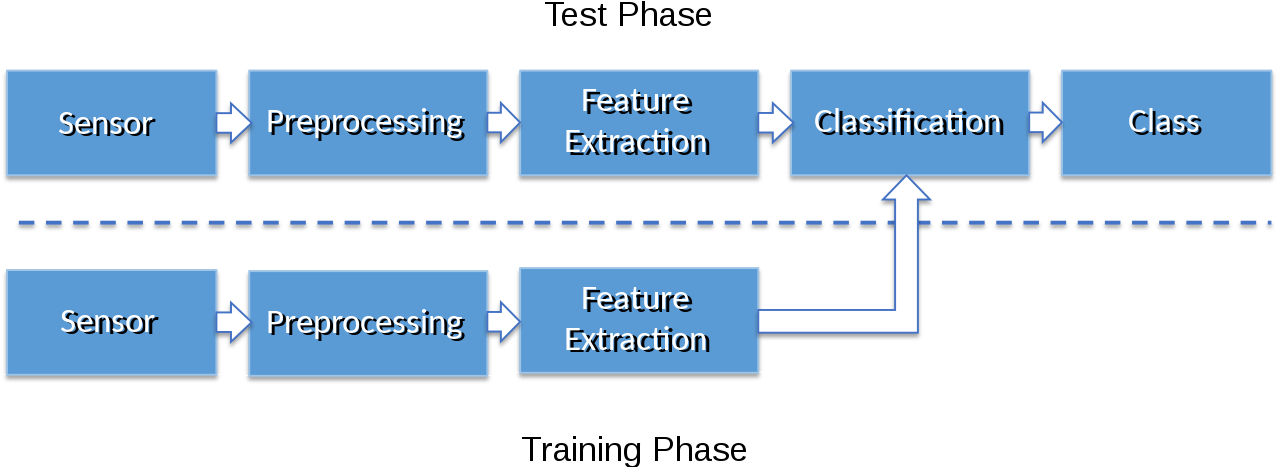}
	\caption{Diagram illustrating the conventional pattern recognition system used for automated decision making. The sensor data is pre-processed and "hand-crafted" features are extracted during both the training and testing phases. In the training phase, a classifier is developed, which is then used in the testing phase to automatically determine the classes (Figure reprinted under CC BY 4.0 \cite{niemann2013pattern}).}
	\label{fig:pattern_rec_pipeline}
\end{figure}

Data-driven modeling is based on the analysis of the data that characterize the system under study. A model is then defined based on the relationships between the state variables (input, internal, and output variables) of the system \cite{solomatine2008data}. Data-driven modeling is essentially implemented with machine learning or pattern recognition methods. Behind both terms is the problem of making automated decisions, for example, distinguishing apples from pears. In the traditional literature \cite{niemann2013pattern}, this process is outlined using the pattern recognition system (cf. fig.~\ref{fig:pattern_rec_pipeline}). Since the internal structure of the model is not known, such a model is also called a black box model.
Within a training phase, the so-called \emph{training data set} is processed and meaningful \emph{features} are extracted. While pre-processing remains in the original data space and involves operations such as noise reduction and image rectification, feature extraction is faced with the task of determining an algorithm that would be able to extract, for example, a distinctive and complete feature representation. This task is usually difficult to generalize. Therefore, it is necessary to redesign such features for each new application essentially.
In the deep learning literature, this approach is often referred to as ``hand-crafted features''. Based on the feature vector $\fex \in \real^n$, the \emph{Classifier} must predict the correct \emph{Class} $\classvar$. This is typically estimated by a function $\prediction = \classifier (\fex)$, which leads directly to the classification result $\prediction$. The parameter vector $\param$ of the classifier is determined during the training phase and later evaluated at an independent \emph{test data set}. A detailed description of this process is given in \cite{maier2019gentle}.

Data-driven modeling eliminates the need to analyze and map already known structures, physical interactions, and similar properties of the system. Instead, the system is described only by the interaction with its environment at the system's inputs and outputs \add{\cite{van2014port,faulwasser2023behavioral}}. Since no knowledge of the internal structure of the system is required to create the model, this method can also be applied in cases where the system structure and the interaction of the components in the system are not or insufficiently known, e.g. in existing systems whose documentation is incomplete. Data-driven models can also be advantageous when the interactions in the system are difficult to describe or parameterize, e.g., in the case of strongly non-linearly coupled state variables. An analysis of the system and a complex parameter identification thus become superfluous \cite{paper}.

\add{Building upon the advantages of data-driven modeling, particularly in scenarios with strongly non-linearly coupled state variables, the Extended Dynamic Mode Decomposition (EDMD) method offers a robust framework for effectively capturing and predicting system behavior \cite{Brunton_Kutz_2019}. This approach leverages the Koopman operator to linearize complex dynamics in an infinite-dimensional space, subsequently approximated by EDMD in a finite-dimensional subspace, enabling linear analysis and control methods to be applied to nonlinear systems. The essence of EDMD in data-driven modeling lies in its ability to use observable functions, such as sensor data, to construct a data-driven surrogate model for predicting system dynamics. Error analysis, such as finite-data error bounds, plays a crucial role in assessing the accuracy and reliability of Koopman-based predictions and controls, providing a probabilistic measure of the approximation and prediction errors \cite{nuske2023finite}. The theoretical foundation of this method is rooted in the Koopman framework, which adeptly manages strong non-linearities by transforming them into a linear, albeit infinite-dimensional, model, thus simplifying analysis and expanding the scope for advanced control strategies, including kernel-based methods \cite{Ding2013Kernel-Based}.}

In data-driven models, an implicit reduction takes place insofar as only those variables that are observable as input or output variables in the interaction with the environment can be considered in the system modeling. The internal structure and the internal variables of the system, on the other hand, are not included in the modeling.

Data-driven models, like analytical models, generally contain internal state variables. However, unlike analytical models, these variables are usually not physical quantities. Rather, the model variables are synthetic quantities, which are therefore not always unambiguously interpretable.

As the most important prior knowledge for the creation of a data-driven model, the determining input and output variables must first be known. In addition, these variables must be observed in all relevant states, whereby the dynamics of the variables must also be represented. The selection of data for typical states used in model building always leads to a limitation of the model created. The learning of the system behavior on the basis of such typical observations implicitly leads to the storage of this behavior as a "good state" of the system, whereby, for example, deviations of the observations from the created model become recognizable.

\subsection{Uncertainty Evaluation}
\add{Black box models, due to their empirical nature, rely heavily on data-driven methods for uncertainty evaluation. Monte Carlo simulations are widely used in this context, providing a means to generate a distribution of possible outcomes based on repeated random sampling.  This approach is particularly valuable for complex systems where analytical methods may not inspire confidence or may be difficult to apply \cite{Basil1999Uncertainty}. Monte Carlo Simulation allows for the assessment of the uncertainty of measurement systems by simulating a range of possible outcomes using random variables. The method is beneficial for evaluating the uncertainty of various measurement systems, including those that cannot be readily addressed by conventional analytical means, due to its ability to take account of partially correlated measurement input uncertainties and its adaptability to complex, nonlinear measurement equations. The key advantage of this approach is its ability to handle non-linear relationships and model interactions without a detailed understanding of the underlying processes. Brando et al. \cite{Brando2020Building} explored the addition of uncertainty scores to black-box predictions, a crucial development for real-world applications where predictive models are often used as black boxes. This work demonstrates the feasibility of quantifying uncertainty in black box models without accessing their internal workings.}

\subsection{Advantages and Limitations}\label{sec:4.2}

An advantage of data-driven modeling is that no explicit physical understanding of the system is required, so modeling can also be performed for complex systems or systems whose internal structure is unknown. By limiting the learning process to observed system states, data-driven models are well suited for the detection of anomalies, i.e. significant deviations of the current system state from a desired or normal state \cite{BlackVSWhite}.

However, data-driven modeling also has limitations and drawbacks. When observing the system, typical inputs and outputs of the system must be available in all relevant forms. If these observations are not available, the model may be incomplete, i.e. the simplification is too strong, or the delineation is not correct. To ensure that the relevant variables are observed and that typical observations and system states are selected in the data selection, a minimum level of system knowledge is required so that expert knowledge cannot be completely dispensed with. A drawback in the plausibility check of models is the aforementioned difficulty in interpreting the model variables \cite{plass2017}. In addition, as mentioned above, data-driven models tend to have comparatively low generalisability, since the restriction of the learning process to an observed operational domain corresponds to a narrowing down to exactly that domain.

However, low generalisability may be desirable: In anomaly detection, the restriction to a certain (error-free) operating range is intended, so that in this case a targeted restriction of the learning process to an observation dataset that is rated as good takes place. Furthermore, data-driven models are usually only slightly abstract. Since the internal structure of the system is not known or used in the modeling, the formation of classes of similar systems is difficult and can only be achieved phenomenologically by analyzing the interactions of the system with its environment.

% Hybrid Modelling

\section{Hybrid modeling: Grey-Box Models}\label{sec:5}

\subsection{Overview}\label{sec:5.1}

Analytical and data-driven modeling are not necessarily in competition with each other, but can rather be combined to exploit the advantages of both approaches. An important reason for this mixing is that the choice of approaches often depends on the professional background of the actors. For example, while engineers are more interested in understanding a system and therefore tend to use analytical models, computer scientists are used to dealing with data and therefore often prefer data-driven models \cite{paper}.

Although in analytical modeling, the physical structure of the system is defined and the model parameters are initially determined by analyzing the system, the quality of the parameter fit can usually be improved by later data-driven optimization. This approach of "data-driven optimized analytical modeling" can be used advantageously, for example, if the internal state variables of the system are known but are coupled by unknown or difficult-to-identify (e.g., strongly non-linear) interactions. In a pure white-box model approach, the unknown interactions would have to be included using probabilistic arguments or as model uncertainties. A combined approach of analytical and data-driven modeling can reduce the overall uncertainty.

If a system modeling is to be essentially data-driven instead, it can still be advantageous to first structurally decompose the overall system into a few, easily identifiable subsystems. Depending on the type of subsystems, this approach of "analytically structured data-driven modeling" can have the advantage that the emerging subsystems can be described more easily in several models that are initially considered separately.

In a recent paper, it was found that the inclusion of known subsystems is generally favorable with regard to the maximum error bounds [Maier 2019b]. The paper shows that the inclusion of prior knowledge in the form of differentiable modules always reduces the maximum error that can be generated by the learning problem. The authors demonstrate the applicability of their theory in grey-box models that incorporate differentiable modules in the trainable CT reconstruction, in the measurement of retinal vessels using hybrids between deep learning and classical image processing, and in the application of image re-binning. The trend towards integrating differentiable known modules into deep networks can thus also be traced back to a solid theoretical basis and shows that classical theory and novel data-driven methods are not in contradiction. Instead, solutions that combine the best of both worlds are preferable. Deep networks can even be reverse-engineered to identify relevant processing modules within the deep network, as shown in \cite{Fu2019}.

Today, we see a steady rise of deep learning models. Obviously, this paper is not able to summarize the totality of deep learning methods. Even focusing only on the domain of medical image analysis would be far beyond the scope of a single paper. However, quite successful attempts to do so can be found in the literature. \add{Ker et al. \cite{ker2018deep} explore the diverse applications of deep learning in medical imaging, demonstrating its ability to improve both diagnostic accuracy and operational efficiency. Litjens et al. \cite{litjens2017survey} provide an in-depth examination of the transformative impact of deep learning on medical imaging, highlighting improvements in accuracy and processing speed. Maier et al. \cite{maier2022known} explore the combination of traditional and deep learning approaches, suggesting innovative ways to advance the field. Despite the comprehensive insights provided by these studies, they do not fully capture the full spectrum of deep learning research in medical imaging. In particular, areas such as image synthesis, multimodal image fusion, and patient-specific modeling and prediction remain less explored in these reviews. This limitation underscores our focused approach, in which we discuss the concepts of grey-box modeling at a high level and illustrate them with selected examples.}

The following sections briefly summarise important developments in the field of grey-box models and their relation to hybrid models. It will first look at the field of Deep Learning from an overarching perspective and show how different approaches succeed in introducing an inductive bias.

A competing approach, as already mentioned by Sutton \add{\cite{sutton1992adapting}}, is meta-learning. Therefore, a brief review of the literature will be provided, summarising relevant methods that are current research at the time of publication of this paper. Finally, we will look at methods that implicitly or explicitly introduce prior knowledge into deep networks and summarise their strengths and weaknesses.

\subsubsection{Deep learning}
Deep learning is generally regarded as an artificial neural network with many layers \cite{lecun2015deep}. However, interpretations of "many" vary widely in the literature from "more than three" \ to thousands of layers. The network that made a major breakthrough in image classification was Alex-Net \cite{krizhevsky2012imagenet}. Alex-Net was then able to approximately halve the error in the image net challenge. In particular, the introduction of specialized operations in networks such as convolutional layers seems to be a key factor for the success of Deep Learning. It seems that the invariances introduced by such layers are beneficial both in terms of parameter reduction and in terms of incorporating \emph{inductive bias} in the network.

While convolutional layers are probably the most popular layer used in Deep Learning, many other invariances can also be encoded in network layers. Bronsteinet al. summarised several important invariances in a recent publication \cite{bronstein2021geometric}. As they demonstrate, lattices, groups, graphs, geodesics and meters provide suitable invariances that lead to certain types of layers, ranging from convolutions to graph layers to recurrent layers. Each of these layers is capable of describing a particular data invariance that can be exploited in a particular mathematical operation. However, there are many other types of layers that use such invariances, such as tensor networks, which also allow the encoding of invariances corresponding to higher-order tensor products. Such techniques are also already used in medical imaging, e.g. in ~image segmentation~\cite{selvan2020tensor}.
A question that currently cannot be answered adequately in the literature is which invariances should be used in which specific order and configuration. A key method to address the problem seems to be the \glqq graduate student descent\grqq method.

\subsubsection{Meta-learning}
Meta-learning attempts to learn the \emph{inductive bias} directly from the data itself. The general approach aims to learn similarities across different tasks. Although a full summary of meta-learning methods is beyond the scope of this paper, only some highlights of the past years are described in this section.

A model-independent approach to meta-learning is presented in \cite{pmlr-v70-finn17a}. Here, the authors propose to separate the meta-task parameters from the task-specific ones. This is achieved by modeling two sets of parameters: one for the generic task and one for the specific task. The developed training strategy produces stable parameters for the meta-task and only a few gradient iterations with few samples allow the adaptation to a specific application.
Several other approaches can be found in the literature, ranging from learning invariances \cite{zhou2021metalearning} to learning prototypical networks \cite{snell2017prototypical} to neural architecture search \cite{zoph2017neural}. However, the differences between the methods are small and vary from dataset to dataset.

Unfortunately, a recent study suggests that simply learning the task-dependent nonlinear distance in the feature space \cite{Sung_2018_CVPR} may prove to be as effective as the above meta-learning approaches without modeling the meta-task at all. The paper shows that common architectures used in Deep Learning are sufficient to accomplish this task.

It is clear that meta-learning has a high potential for future applications in modeling inductive bias. However, none of the methods found in the literature can demonstrate this for practical applications while outperforming current deep learning methods.

\subsubsection{Pre-knowledge as regularisation}
Pre-knowledge as regularisation is an approach that attempts to incorporate prior knowledge into deep learning models. The most common approach is to use the loss function to embed prior knowledge using a regularisation term \cite{willard2020integrating}. As such, the loss is extended by an additive part that penalizes the model if it deviates from a priori knowledge given as an analytical model during training. A major drawback of this approach is that the deviation from the physical model is only penalized during training, and there is no mechanism to actually guarantee the plausibility of the model during the testing period.
Another very common approach to making deep learning models plausible is to embed them in a constrained learning framework. This is often done through reinforcement learning or imitation learning. This approach has proven to be very effective in games, as the rules of the game can be used and sophisticated search algorithms such as Monte Carlo tree search \cite{silver2016mastering} can be embedded in the machine learning algorithms. This approach is also very popular for medical imaging applications, as shown in \cite{ghesu2017multi}. Unfortunately, the reinforcement learning setup is computationally expensive during training, and many events of the present learning task have to be repeated to train a good network.

\subsubsection{Known operator learning}
Known operator learning\add{, a term introduced in earlier work,} is an approach to embed analytical models directly into deep networks already known from classical theory. It is based on the assumption that the function to be learned can be decomposed into modules, some of which are known and some of which have to be learned from data. \add{The term's specific use in our context aims to highlight this unique integration, distinguishing it from the broader, more general use of "operator learning" in a functional-analytic sense.} 

\add{Deep networks are often observed to organize themselves into such modular configurations, supporting the viability of this approach \cite{filan2020pruned}. Incorporating known operations into the network architecture not only reduces the maximum error bound \cite{maier2019learning} but also decreases the number of parameters that need to be trained. This reduction in trainable parameters can lead to a smaller required training dataset. Moreover, this method has been shown to enhance the generalization capabilities of the network, a benefit that has been empirically verified \cite{syben2018deriving}.}

\begin{figure}[tb]
	\centering
	\includegraphics[width=1\linewidth]{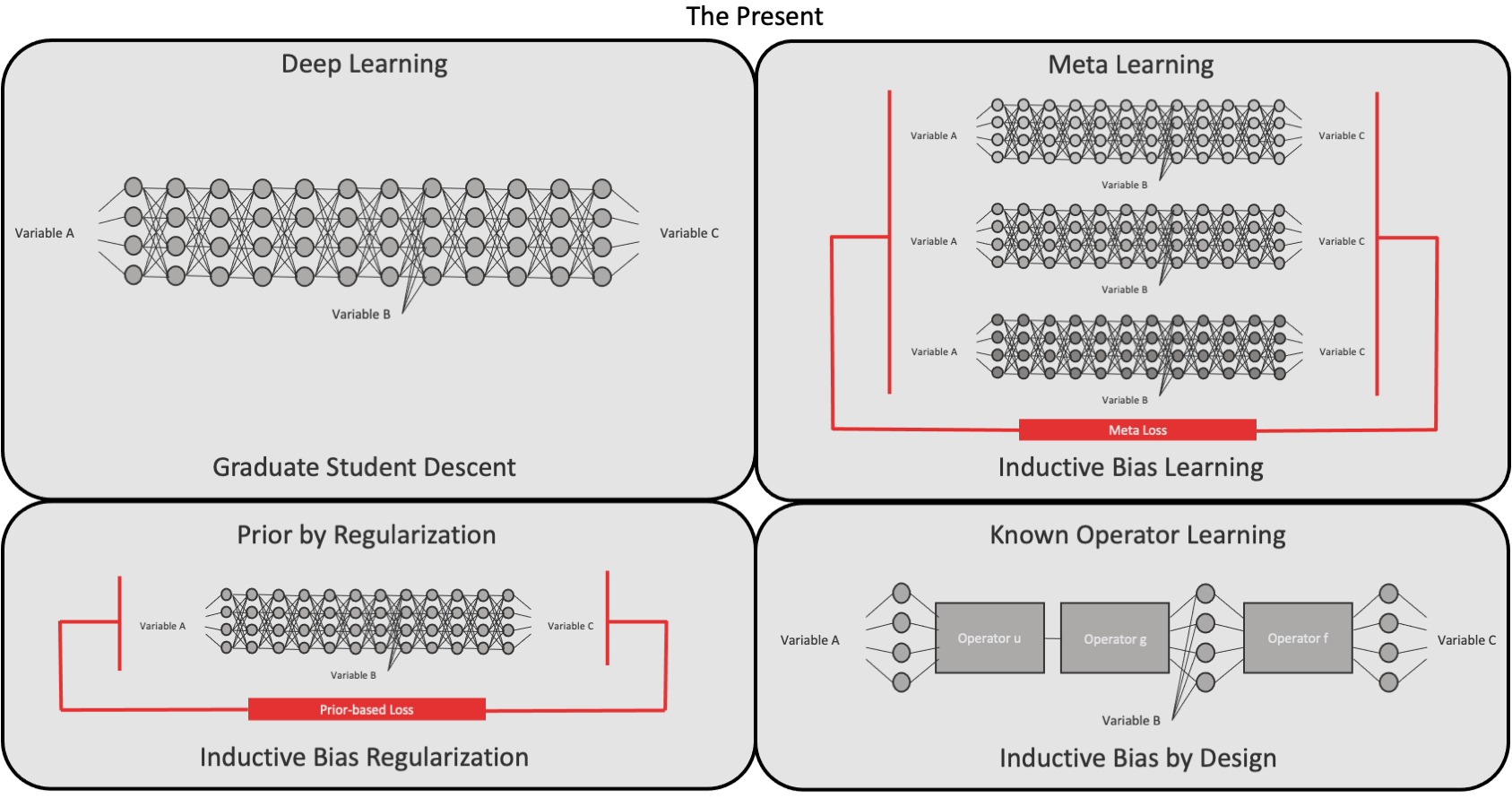}
	\caption{Today, several methods are commonly used to introduce an \emph{inductive bias} into machine learning models. They range from "graduate student descent" in Deep Learning to Meta Learning and regularisation methods. The image is reprinted under CC BY 4.0 \cite{maier2022known}.}
	\label{fig:thepresent}
\end{figure}

\subsubsection{Physics-Informed Learning}
\add{ Physics-informed learning, particularly through models like Physics-Informed Neural Networks (PINNs), incorporates physical laws, often in the form of differential equations, directly into the learning process \cite{karniadakis2021physics}. This approach ensures that the learned models adhere to known physical principles, which is particularly useful when data is scarce or noisy. For instance, Physics-Informed Neural Operators (PINOs) utilize both data and physics constraints to learn the solution operator of parametric Partial Differential Equations (PDEs), offering a hybrid solution that overcomes the limitations of purely data-driven methods and purely physics-based approaches. PINOs can incorporate data and PDE constraints at different resolutions, allowing for the combination of coarse-resolution data with higher-resolution PDE constraints without loss in accuracy, a property known as discretization-invariance. This enables efficient learning even in scenarios where no data is available, by optimizing PDE constraints across multiple instances rather than a single instance, as seen in other approaches like PINNs \cite{Li2021Physics-Informed}.
}
\add{ A recent paper \cite{li2021kohn} showed that learning the entire hydrogen dissociation curve is possible and experimentally validated the generalization properties predicted by the well-known operator learning theory. The approach is also suitable for extracting symbolic equations using graph networks, as shown in \cite{cranmer2019learning}.
}

\subsection{Uncertainty Evaluation}
\add{In grey box modeling, uncertainty evaluation often involves statistical methods to estimate the uncertainty associated with empirically derived model parameters. This approach effectively bridges the gap between theoretical models and empirical data, making it particularly useful in systems lacking complete theoretical understanding. Bayesian inference is a common technique used in this context, allowing for the incorporation of prior knowledge about the parameters into the uncertainty evaluation process \cite{Cheng2018Analysis}. This approach is based on Bayes' theorem, which updates the probability for a hypothesis as more evidence or information becomes available. It provides a principled way to combine new data with prior beliefs, and is particularly useful in complex systems where the true state may not be directly observable. For example, Bayesian methods have been applied to the evaluation of measurement uncertainty, where they can fully integrate prior and current sample information, determine the prior distribution based on historical data, and deduce the posterior distribution by integrating the prior distribution with current sample data using the Bayesian model. This enables the optimization estimation of uncertainty, reflecting the latest information on the accuracy of the measurement system. Boumans discussed a more objective Type B evaluation achieved through model-based uncertainty evaluations involving grey-box modeling and validation, highlighting the application of grey-box approaches in metrology \cite{Boumans2013Model-based}.}

\subsection{Advantages and Limitations}
\add{Hybrid modeling offers a balanced approach by integrating both data-driven elements and physical knowledge of the system. This dual approach enables the modeling of complex systems where some aspects of the internal structure may be understood, while others might be too intricate or unknown. By incorporating physical laws or system dynamics into the models, grey-box approaches can enhance interpretability and reliability, particularly in extrapolating beyond the range of observed data.}

\add{Unlike purely data-driven models that rely entirely on observed states and can struggle with generalizability and interpretability, grey-box models leverage known system equations or behaviors to structure the model. This can significantly improve the model's ability to generalize from limited data by grounding the predictions in established physical principles. For instance, in a mechanical system, equations of motion could guide the model structure, ensuring that predictions adhere to fundamental laws like conservation of energy or momentum.}

\add{However, grey-box modeling also requires careful consideration. The model's effectiveness hinges on the accuracy and relevance of the incorporated physical knowledge. Incorrect assumptions about the system's dynamics can lead to models that are no more reliable than their purely data-driven counterparts. Moreover, the process of integrating physical knowledge into the model requires a deep understanding of the system, which might not always be available or might necessitate significant expertise.
}

\add{In metrology, the application of Grey-Box Models enhances the precision and reliability of measurements in complex systems where direct quantification can be challenging. By integrating physical principles with empirical data, these models are instrumental in calibrating measurement instruments, dynamically compensating for environmental influences, and quantifying measurement uncertainties more accurately. For instance, in scenarios where environmental conditions such as temperature and humidity affect measurement accuracy, Grey-Box Models can adjust measurements in real-time, ensuring consistent accuracy. Additionally, they prove invaluable in non-invasive measurement techniques, allowing for the estimation of inaccessible quantities through indirect measurements and known physical relationships. This hybrid approach not only improves the direct application of metrological instruments but also aids in proactive maintenance by predicting performance degradation due to wear and tear, thereby maintaining the integrity of precision measurements. While offering significant advantages in enhancing measurement precision and reliability, they also present limitations, particularly in their dependency on the accuracy of the embedded physical models. If the incorporated physical principles do not fully capture the system's dynamics or if they are based on incorrect assumptions, the model's predictions can be misleading, potentially compromising measurement accuracy. This limitation underscores the importance of comprehensive system understanding and the need for meticulous validation of the physical models used.  Grey-Box models, therefore, enhance metrology by improving measurement capabilities and reliability, but their effectiveness hinges on the accuracy of the integrated physical models to ensure measurement integrity.}

% Intelligent, adaptive modelling approaches

\section{Intelligent Adaptive Systems}\label{sec:6}

\add{Intelligent adaptive systems refer to advanced technological frameworks that autonomously adjust and optimize their operations in response to varying environmental conditions and user requirements. These systems combine artificial intelligence, machine learning, sensors, actuators, and control logic to create dynamic solutions that can learn from and adapt to new situations without human intervention. In the context of metrology, intelligent adaptive systems play a transformative role.}

\add{Intelligent adaptive systems in metrology involve the integration of sensors, actuators, and control logic into measurement systems to enhance their adaptability, precision, and responsiveness. These systems can dynamically adjust their behavior or operation in response to changes in the environment or measurement conditions, improving the accuracy and reliability of metrological processes. For instance, in aerospace metrology, intelligent structures equipped with sensors and actuators can modify their mechanical states or characteristics—like position, velocity, stiffness, or damping—to maintain measurement accuracy under different operational conditions \cite{Crawley1994Intelligent}.
}

\add{This section explores the transformative impact of intelligent adaptive systems and digital twins in metrology, highlighting their integration of AI, machine learning, and real-time sensor data to improve measurement accuracy and adaptability. It explores the seamless fusion of physical and digital entities through digital twins, enabling a continuous feedback loop that refines metrological practices. It also outlines advances in intelligent learning systems, such as the artificial neural twin and reservoir computing, and their role in improving predictive accuracy and system adaptability. Using examples such as soft sensors in bioprocess control and the Frangi filter in advanced imaging, this section illustrates the multiple benefits and challenges of these innovative technologies in metrology.}

\subsection{Intelligent Adaptive and Autonomous Systems}\label{sec:6.1}
\add{In the specialized domain of Intelligent Adaptive and Autonomous Systems, the adaptive capabilities of intelligent systems are extended to include autonomy in decision making and task execution. These systems are self-sufficient, using advanced AI to react to changes and independently handle complex situations. After learning, these systems can be either static, optimized for a single, predefined task, or flexible, capable of generalizing their learned knowledge to a variety of related challenges. Incorporating principles of safe and continual learning further ensures that these adaptations occur without compromising system integrity or performance over time, thereby broadening their applicability and effectiveness in dynamic settings \cite{garcia2015comprehensive}. To illustrate this, we take a closer look at two examples here.}

\add{The first example discusses the use of soft sensors in industrial processes. Soft sensors, also known as virtual sensors, are a type of machine learning that aims to infer unmeasurable process variables from available sensor data. These algorithms are designed to learn meaningful representations from operational data, often without requiring explicit human-crafted features or extensive domain knowledge. We use the example of applying soft sensors to bioprocess control, specifically by applying a soft sensor for sequential filtering of metabolic heat signals \cite{Paulsson2014}. The approach uses temperature sensor signals from a bioreactor's cooling system to estimate a microbial culture's metabolic heat. This estimation allows for the derivation of specific growth rates and active biomass concentrations, which can enhance bioprocess monitoring and control capabilities. The robustness of the soft sensor is significantly enhanced by implementing sequential digital signal filtering. This improvement makes it suitable for industrial applications where cultures generate low metabolic heat in environments with high noise levels. }

\begin{figure}
\centering
\includegraphics[width=0.9\linewidth]{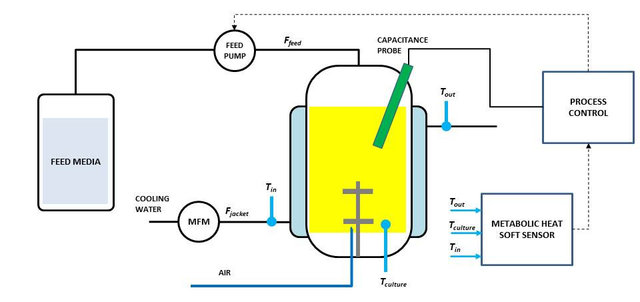}
\caption{Illustration of the bioprocess control using soft sensors based on sequential filtering of metabolic heat signals \cite{Paulsson2014}.}
\label{fig:bioprocesscontrol}
\end{figure}

\add{This method relies on a data-driven structure, which is beneficial for the specific task of bioprocess control. However, the representations learned are highly specialized, making applying them directly to unrelated tasks almost impossible without substantial retraining or adaptation. This underscores the importance of customized approaches to the application of soft sensors in specific domains.
}

\add{The situation is different when using a trained Frangi filter \cite{Fu2017}. The adaptation of the Frangi filter into a trained neural network model has notably advanced the automatic segmentation of the retinal vascular tree, a task complicated by the intricate nature of retinal vessels and often compromised image quality. Originating from Frangi et al.'s \cite{Frangi1998} vascular enhancement methodology, which relies on the eigenvalue analysis of the Hessian matrix at various Gaussian scales, the trained Frangi filter incorporates the strengths of neural networks to refine segmentation accuracy. This hybrid approach not only preserves the inherent vessel-identifying capabilities of the original Frangi filter but also capitalizes on the adaptive learning potential of neural networks, resulting in significant improvements in identifying vessels of varying sizes in retinal images.}

\begin{figure}[tb]%
	\centering
	\subfloat[][Before \label{fig:vessel:a}]{\includegraphics[width=0.4\linewidth]{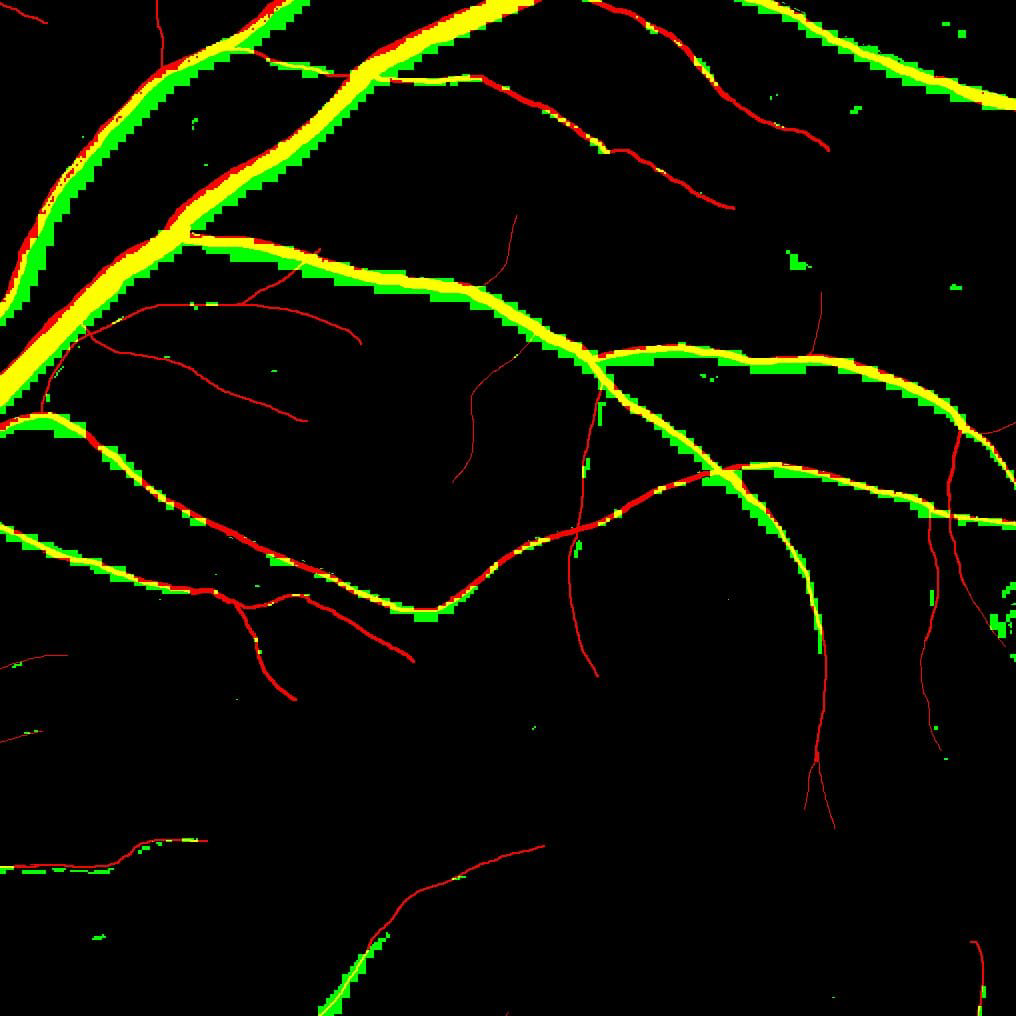}}%
	\qquad
	\subfloat[][After\label{fig:vessel:b}]{\includegraphics[width=0.4\linewidth]{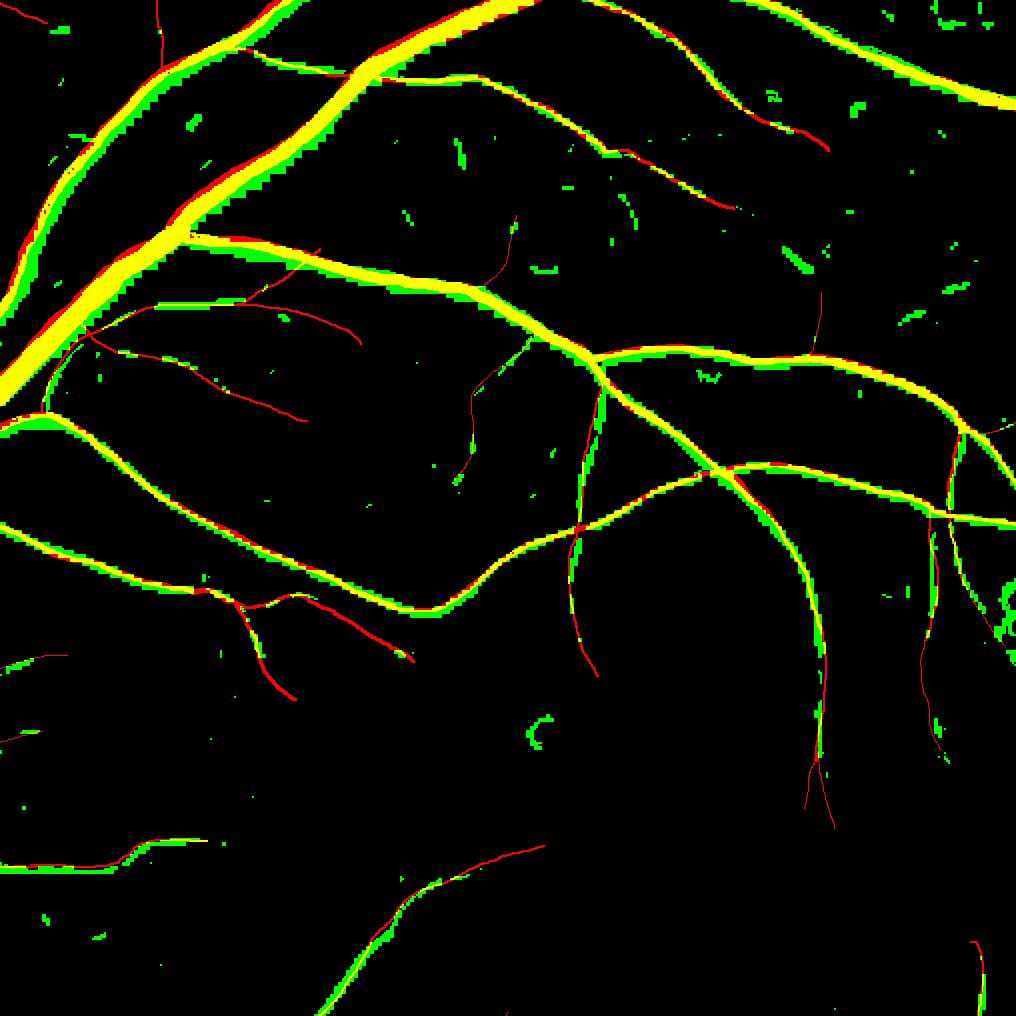}}%
\caption{Results of the vessel segmentation before and after training. The colors red, green, and yellow represent the manual annotations, the segmentation results, and the overlaps between them, respectively \cite{Fu2017}.}%
\label{fig:vessel}
\end{figure}

\add{Expanding its utility beyond ophthalmic imaging, the Frangi filter has also been adapted for segmenting fat and fascia in ultrasound images, as demonstrated in Rybakov et al.'s study \cite{Rybakov2018}. This application underscores the filter's versatility, effectively identifying fascia—which appears similar to white vessels on ultrasound images—and fat as a thick, dark, tubular structure, thereby streamlining the segmentation process for biological research and clinical diagnostics.}

\add{Furthermore, the versatility of the Frangi filter extends beyond medical imaging. Its application has proven effective in the automation of book page extraction from high-resolution 3D scans \cite{Stromer2018}. Here the biggest challenge lies in accurately distinguishing and separating the individual pages within the tightly compressed and highly detailed 3D volume.} 

\add{In the domain of metrology, the Frangi filter finds a novel application in the enhancement of vessel-like structures within optoacoustic imaging, a technique that merges optical and acoustic imaging to provide detailed insights into the properties of vascular structures. A pivotal study by Longo et al. \cite{Longo2020Assessment} rigorously assessed the Hessian-based Frangi vesselness filter's efficacy in optoacoustic images, validating its capability to accentuate vessel-like forms in phantom models and further examining its performance in vivo with the aid of gold nanorods. This research illuminated the significant impact of factors such as contrast, filter scales, and angular tomographic coverage on image quality, while also addressing the potential for artifact generation that could lead to misinterpretations. The findings offer valuable recommendations for the judicious application of the Frangi filter and similar vesselness filters in optoacoustic imaging, ensuring the accuracy of measurements and interpretations crucial in the metrological evaluation of vascular structures \cite{Longo2020Assessment}.}

\add{The Frangi filter is a compelling example of how intelligent adaptive and autonomous systems can evolve. It demonstrates the potential for these systems to perform predefined tasks with increased efficiency and apply their capabilities in novel, dynamic environments, expanding the horizons of what autonomous systems can achieve in the realm of intelligent adaptation.}

\subsection{Role and Potential \add{of} Digital Twins} \label{sec:6.2}
\add{In the field of metrology, precise measurements and analysis are of vital importance. The concept of digital twins represents an important innovation as it enables the accurate reproduction of an object's condition, behavior, and changes in real-time. Digital twins create a digital counterpart to a physical entity. This continuous synchronization throughout an object's lifecycle enables deep integration of physical measurements with digital simulations, improving metrology applications with high accuracy and efficiency.}

\add{The foundation of digital twins lies in the idea that a physical object and its digital representation can be interconnected, allowing for real-time data transmission and analysis. This concept is not tied to a specific technology but is a versatile framework that can be realized through various advanced technological solutions \cite{Liu2020}.} A central problem that the digital twin model should solve is the contradiction between the simplified virtual model and the complex behavior of the physical object. The origin of the Digital Twin is attributed to Michael Grieves and his work with John Vickers of NASA, with Grieves introducing the concept in 2003 in a lecture on product lifecycle management \cite{Grieves2015}. 
The original description defines a Digital Twin as a virtual representation of a physical product that contains information about the product. In an early paper \cite{Grieves2015} Grieves expanded this definition and described the Digital Twin as consisting of a physical product made up of three components, a virtual representation of that product and the bidirectional data links that transfer data from the physical to the virtual representation and information and processes from the virtual representation to the physical product.
Grieves described this flow as a cycle between the physical and virtual states (mirroring or twinning).
\begin{figure}[tb]
	\centering
	\begin{tikzpicture}
		\tikzstyle{none}=[]
		\tikzstyle{square node}=[draw=black, shape=rectangle,thick, tikzit category=nodes, minimum height = 1.3cm, minimum width = 3.5cm]
		\tikzstyle{dashed arrow}=[->, dashed, draw = gray, thick]
		\tikzstyle{gray dashed line}=[-, draw=gray, dashed, thick]
		\tikzstyle{gray pointer}=[->, draw=gray, thick]
		\tikzstyle{gray line}=[-, draw=gray, thick]
		\begin{pgfonlayer}{nodelayer}
			\node [style=square node] (0) at (0, 1) {Virtual Space};
			\node [style=square node] (1) at (0, -1) {Real Space};
			\node [style=none] (2) at (4, 1) {};
			\node [style=none] (3) at (4, -1) {};
			\node [style=none] (4) at (-4, 1) {};
			\node [style=none] (5) at (-4, -1) {};
			\node [style=none, text width=3cm] (6) at (5.75, 0) {Information/ Processes};
			\node [style=none] (7) at (-4.75, 0) {Data};
		\end{pgfonlayer}
		\begin{pgfonlayer}{edgelayer}
			\draw [style=gray line](1) to (5.center);
			\draw [style=gray line](5.center) to (4.center);
			\draw [style=gray pointer] (4.center) to (0);
			\draw [style=gray dashed line] (0) to (2.center);
			\draw [style=gray dashed line] (2.center) to (3.center);
			\draw [style=dashed arrow] (3.center) to (1);
		\end{pgfonlayer}
	\end{tikzpicture}
	\caption{Reflection or twinning between the physical and virtual spaces \cite{Jones2020}.}
	\label{fig:mirrortwinning}
\end{figure}
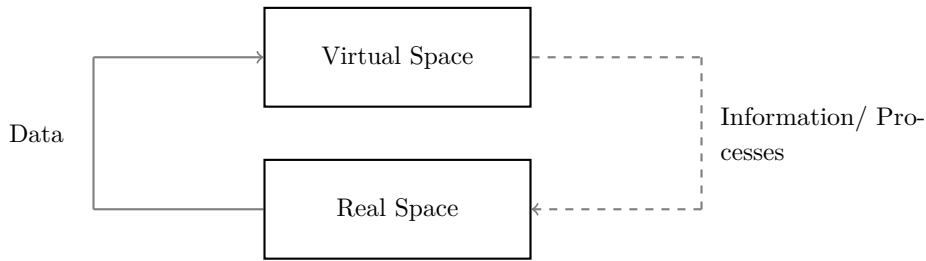
Since the inception of the Digital Twin in 2003, the concept has gained interest and has been listed by Gartner as a key strategic technology trend for 2019 \cite{Jones2020}. This growth is largely driven by advances in related technologies and initiatives such as Internet-of-Things, Big Data, multi-physical simulation and Industry 4.0. In addition, there are real-time sensors and sensor networks, data management, data processing and the drive towards a data-driven and digital future of manufacturing. 

Data is the foundation of digital twins. Sensors, measuring devices, RFID tags and readers, cameras, scanners, etc. are selected and integrated to collect data for the digital twin. The data is then to be transmitted in real-time or near real-time. However, such data is usually very large and diverse, making it difficult and costly to transfer to the digital twin on, for example, a cloud server. Therefore, edge computing is an ideal method for pre-processing the collected data to reduce the network load and eliminate the risk of data loss \cite{Jones2020}.\

The model is the core of the digital twin. Digital twin models consist of semantic data models and physical models. Semantic data models are trained on known inputs and outputs using AI methods. Physical models require a comprehensive understanding of physical properties and their mutual interaction. Therefore, multi-physical modeling is essential for the realistic modeling of the digital twin. Simulation is an important aspect of the digital twin in this regard. Digital twin simulation allows the virtual model to interact with the physical entity in real-time.
The structure of digital twins described below is based on the results of the literature review by Jones et al. \cite{Jones2020}.

\subsubsection{Construction of a Digital Twin}

In general, the digital twin consists of a physical entity, a \add{virtual} counterpart and the data connections in between. In total, digital twins can be divided into 12 components. First, a distinction can be made between two environments, the physical environment and the virtual environment. The physical environment refers to the "real" space in which the physical entity is located. Aspects of these environments are measured and transferred to the virtual twin to ensure an accurate virtual environment upon which simulations, optimizations, and/or decisions are made. Physical processes refer to the activities performed by the physical entity in the physical environment.
The virtual environment exists in the digital domain and is a mirror of the physical environment, where twinning is achieved through physical measurement technology (i.e. sensors), transferring important measurements from the physical to the virtual environment. Virtual processes refer to the activities performed using the virtual entity within the virtual environment. The term parameters in this context refers to the types of data, information and processes that are passed between the physical and virtual twins. 

The physical-virtual links are the means by which the state of the physical entity is transferred to and realized in the virtual environment - i.e., updating the virtual parameters so that they reflect the values of the physical parameters. These include Internet-of-Things sensors, web services, 5G, and other customer requirements \cite{Jones2020}. The connection itself consists of a measurement phase, in which the state of the physical entity is recorded, and a realization phase, in which the delta between the physical and the digital entity is determined, and the virtual entity is updated accordingly. For example, the temperature change of a physical engine is measured with an Internet-of-Things thermometer (measurement phase), the temperature measurement is transmitted to the virtual environment via a web service, and a virtual process determines the temperature difference between the physical engine and the virtual engine and updates the virtual engine so that both measurements are equal (realization phase). This continuous connection between the physical and virtual engine is a distinguishing feature between the Digital Twin and more traditional simulation and modeling approaches, where analysis is often performed "off-line". The physical-virtual link allows monitoring of state changes that occur both in response to conditions in the physical environment and state changes that occur in response to interventions by the digital twin itself. Thus, if a change in engine speed were to be made as a result of temperature measurements, the physical-digital link would also measure the effects of this intervention. Grieves describes the virtual-physical connection as the flow of information and processes from the virtual to the physical, i.e. the digital twin contains the functionality to realize a change in the physical state physically.

The link from virtual to physical environment mirrors the link from physical to virtual environment in that it also contains measurement as well as realization phases. Virtual processes and measurement methods determine and measure an optimal set of parameter values within a physical entity or environment, and realization methods determine the delta between these new values and the state and update the state of the physical entity accordingly. For example, in response to an increased engine temperature that exceeds a certain threshold, the effect of engine speed can be modeled. This allows a speed to be calculated that sufficiently reduces the temperature, thus adjusting the physical engine speed.
\begin{figure}[tb]
	\centering
	\begin{tikzpicture}[scale=0.37,  every node/.style={scale=0.7}]
		% Node styles
		\tikzstyle{none}=[]
		\tikzstyle{rotated clockwise node}=[fill=white, tikzit category=nodes, rotate = -90]
		\tikzstyle{rotated counterclockwise node}=[fill=white, tikzit category=nodes, rotate = 90]
		\tikzstyle{red node}=[fill=red, tikzit category=nodes, shape=circle, draw=black]
		\tikzstyle{blue node}=[fill=blue, shape=circle, draw=black, tikzit category=nodes]
		\tikzstyle{green node}=[tikzit fill=green, fill=green, shape=circle, draw=black, tikzit category=nodes]
		\tikzstyle{yellow square}=[draw=black, fill=yellow, shape=rectangle]
		\tikzstyle{dotted square}=[draw=blue, fill=white, shape=rectangle, dotted]
		\tikzstyle{square node}=[draw=black, shape=rectangle, tikzit category=nodes]
		\tikzstyle{circle node}=[draw=black, shape=circle, tikzit category=nodes]
		\tikzstyle{blue node 2}=[fill={rgb,255: red,128; green,0; blue,128}, draw=black, shape=circle, tikzit fill=blue]
		\tikzstyle{ellipse node}=[draw = black, shape=ellipse, minimum size=10pt, align=center, tikzit category=nodes]
		
		% Edge styles
		\tikzstyle{dashed edge}=[<->, dashed]
		\tikzstyle{dashed arrow}=[->, dashed]
		\tikzstyle{blue pointer}=[->, draw=blue]
		\tikzstyle{blue dashed pointer}=[->, draw=blue, dashed]
		\tikzstyle{blue dashed line}=[-, draw=blue, dashed]
		\tikzstyle{cyan dashed line}=[-, draw=cyan, dashed]
		\tikzstyle{gray dashed line}=[-, draw=gray, dashed]
		\tikzstyle{pointer}=[->, draw=black]
		\tikzstyle{thick grey arrow}=[->, draw=gray, thick]
		\begin{pgfonlayer}{nodelayer}
			\node [style=ellipse node] (0) at (-5.5, 5.5) {Virtual Unit};
			\node [style=ellipse node] (1) at (5.5, 5.5) {Virtual Unit};
			\node [style=ellipse node] (2) at (5.5, -5.5) {Physical Unit};
			\node [style=ellipse node] (3) at (-5.5, -5.5) {Physical Unit};
			\node [style=square node] (4) at (12, 2.5) {Measurement};
			\node [style=square node] (5) at (12, -2.5) {Realization};
			\node [style=square node] (6) at (-12, 2.5) {Realization};
			\node [style=square node] (7) at (-12, -2.5) {Measurement};
			\node [style=none] (8) at (-5.5, 6.75) {\color{cyan}Virtual Process};
			\node [style=none] (9) at (5.5, 6.75) {\color{cyan}Virtual Process};
			\node [style=none] (10) at (-5.5, -6.75) {\color{cyan}Physical Process};
			\node [style=none] (11) at (5.5, -6.75) {\color{cyan}Physical Process};
			\node [style=none] (12) at (-11.5, 0) {};
			\node [style=none] (13) at (11.5, 0) {};
			\node [style=none] (14) at (0, 1.25) {Virtual Environment};
			\node [style=none] (15) at (0, -1.25) {Physical Environment};
			\node [style=none] (16) at (-9.5, 8) {};
			\node [style=none] (17) at (-9.5, 4) {};
			\node [style=none] (18) at (-1.5, 4) {};
			\node [style=none] (19) at (-1.5, 8) {};
			\node [style=none] (20) at (1.5, 8) {};
			\node [style=none] (21) at (1.5, 4) {};
			\node [style=none] (22) at (9.5, 4) {};
			\node [style=none] (23) at (9.5, 8) {};
			\node [style=none] (24) at (-9.75, -4) {};
			\node [style=none] (25) at (-1.25, -4) {};
			\node [style=none] (26) at (1.25, -4) {};
			\node [style=none] (27) at (9.75, -4) {};
			\node [style=none] (28) at (-1.25, -7.75) {};
			\node [style=none] (29) at (1.25, -7.75) {};
			\node [style=none] (30) at (-9.75, -7.75) {};
			\node [style=none] (31) at (9.75, -7.75) {};
			\node [style=rotated counterclockwise node] (34) at (-16, 0) {\color{blue}\small{Physical-virtual connection/Twinning}};
			\node [style=rotated clockwise node] (35) at (16, 0) {\color{blue}\small{Virtual-physical connection/Twinning}};
			\node [style=none] (37) at (15.25, -5.5) {};
			\node [style=none] (38) at (11, -5.5) {};
			\node [style=none] (39) at (15.25, 0) {};
			\node [style=none] (40) at (15.25, 5.5) {};
			\node [style=none] (41) at (11, 5.5) {};
			\node [style=none] (42) at (12.25, 0) {};
			\node [style=none] (43) at (15, 0) {};
			\node [style=none] (44) at (-15.25, -5.5) {};
			\node [style=none] (45) at (-11, -5.5) {};
			\node [style=none] (46) at (-15.25, 0) {};
			\node [style=none] (47) at (-15.25, 5.5) {};
			\node [style=none] (48) at (-11, 5.5) {};
			\node [style=none] (49) at (-12.25, 0) {};
			\node [style=none] (50) at (-15, 0) {};
			\node [style=none] (51) at (-17.5, 10) {};
			\node [style=none] (52) at (18, 10) {};
			\node [style=none] (53) at (-17.75, -9.5) {};
			\node [style=none] (54) at (18, -9.5) {};
			\node [style=none] (55) at (-16.5, 0) {};
			\node [style=none] (56) at (-17.5, 0) {};
			\node [style=none] (57) at (16.5, 0) {};
			\node [style=none] (58) at (17.75, 0) {};
			\node [style=none] (59) at (0, 10) {};
			\node [style=none] (60) at (0, -9.5) {};
			\node [style=none] (61) at (-17.75, -1) {};
			\node [style=none] (62) at (18, 0.5) {};
			\node [style=none] (63) at (11.5, -9) {};
			\node [style=none] (64) at (11.75, -10) {};
			\node [style=none] (65) at (12.5, -9.5) {};
			\node [style=none] (66) at (10.75, -9.5) {};
			\node [style=none] (67) at (11.75, -10.75) {\color{blue}\small{Twinning Rate}};
			\node [style=none] (68) at (0, 10.75) {\color{blue}\small{Twinning}};
			\node [style=dotted square] (69) at (-12, -9.5) {\color{blue}\scriptsize{\begin{tabular}{cc}
						Parameter 1: & Value \\
						Parameter 2: & Value \\ \dots& \dots \\
						Parameter n: & Value \\
						
			\end{tabular}}};
		\end{pgfonlayer}
\begin{pgfonlayer}{edgelayer}
			\draw [style=thick grey arrow, in=180, out=90, looseness=1.25] (6) to (0);
			\draw [style=thick grey arrow, in=90, out=0, looseness=1.25] (1) to (4);
			\draw [style=thick grey arrow] (4) to (5);
			\draw [style=thick grey arrow, in=0, out=-90, looseness=1.25] (5) to (2);
			\draw [style=thick grey arrow, in=630, out=180, looseness=1.25] (3) to (7);
			\draw [style=thick grey arrow] (7) to (6);
			\draw [style=blue dashed line] (41.center) to (40.center);
			\draw [style=blue dashed line] (39.center) to (37.center);
			\draw [style=blue dashed pointer] (40.center) to (39.center);
			\draw [style=blue dashed pointer] (37.center) to (38.center);
			\draw [style=blue dashed line] (45.center) to (44.center);
			\draw [style=blue dashed pointer] (44.center) to (46.center);
			\draw [style=blue dashed pointer] (47.center) to (48.center);
			\draw [style=blue dashed line] (46.center) to (47.center);
			\draw [style=blue dashed line] (63.center) to (64.center);
			\draw [style=blue dashed line] (64.center) to (65.center);
			\draw [style=blue dashed line] (66.center) to (63.center);
			\draw [style=blue dashed line] (65.center) to (54.center);
			\draw [style=blue dashed pointer] (66.center) to (60.center);
			\draw [style=blue dashed pointer] (52.center) to (62.center);
			\draw [style=blue dashed pointer] (51.center) to (59.center);
			\draw [style=blue dashed pointer] (53.center) to (61.center);
			\draw [style=blue dashed line] (61.center) to (51.center);
			\draw [style=blue dashed line] (59.center) to (52.center);
			\draw [style=blue dashed line] (62.center) to (54.center);
			\draw [style=cyan dashed line] (16.center) to (19.center);
			\draw [style=cyan dashed line] (19.center) to (18.center);
			\draw [style=cyan dashed line] (17.center) to (18.center);
			\draw [style=cyan dashed line] (16.center) to (17.center);
			\draw [style=cyan dashed line] (20.center) to (23.center);
			\draw [style=cyan dashed line] (23.center) to (22.center);
			\draw [style=cyan dashed line] (22.center) to (21.center);
			\draw [style=cyan dashed line] (21.center) to (20.center);
			\draw [style=cyan dashed line] (24.center) to (25.center);
			\draw [style=cyan dashed line] (25.center) to (28.center);
			\draw [style=cyan dashed line] (28.center) to (30.center);
			\draw [style=cyan dashed line] (30.center) to (24.center);
			\draw [style=cyan dashed line] (26.center) to (27.center);
			\draw [style=cyan dashed line] (27.center) to (31.center);
			\draw [style=cyan dashed line] (31.center) to (29.center);
			\draw [style=cyan dashed line] (29.center) to (26.center);
			\draw [style=gray dashed line] (12.center) to (13.center);
			\draw [style=gray dashed line] (42.center) to (39.center);
			\draw [style=gray dashed line] (35) to (58.center);
			\draw [style=gray dashed line] (50.center) to (49.center);
			\draw [style=gray dashed line] (56.center) to (55.center);
			\draw [style=blue dashed line] (60.center) to (69);
			\draw [style=blue dashed line] (53.center) to (69);
		\end{pgfonlayer}
	\end{tikzpicture}
	\caption{The physical-virtual and virtual-physical twinning process \cite{Jones2020}.}
	\label{fig:twinningprocess}
\end{figure}
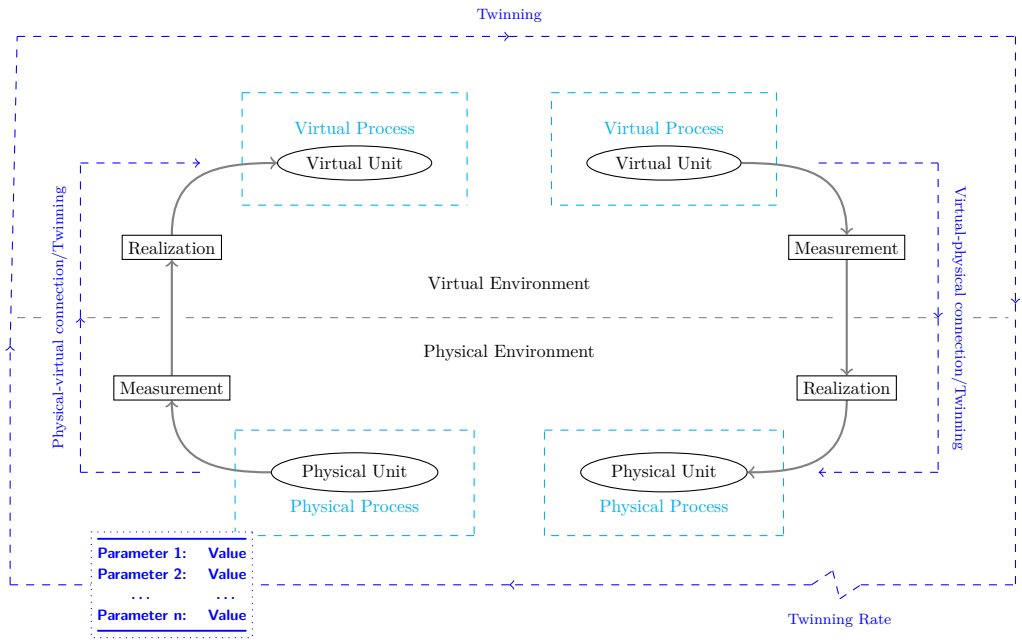\

Twinning describes the synchronization of the virtual and physical states, e.g., measuring the state of a physical entity and translating this state into the virtual environment so that the virtual and physical states are "the same," i.e., all virtual parameters have the same value as the physical parameters. A change that occurs in either the physical or virtual entity is measured before it is realized to the corresponding virtual/physical twin. If both states are the same, the units are " twinned ". The twinning rate is then the frequency at which twinning occurs. 

In Fig. \ref{fig:twinningprocess} it is shown how physical/virtual processes act on the corresponding physical/virtual entity, these processes causing a change in the state of this entity via its parameters. This change of state is captured by measurement methods, transmitted via physical-virtual and virtual-physical connections, and realized in the other (virtual/physical) environment by synchronizing all parameters. Both the virtual and physical environments contain the means to measure and realize changes of state. The process (change → measurement →, realization) is the twinning process and runs in both directions, from virtual to physical and from physical to virtual.

\subsection{New Directions for Intelligent Learning Systems} \label{sec:5.3}

\add{In the rapidly changing field of Intelligent Autonomous Systems, particularly in metrology, new developments are constantly emerging, reshaping the way these systems learn and adapt. This section aims to highlight two significant advancements in this field. The first is the concept of the Artificial Neural Twin, an innovative extension of digital twins that integrates neural network capabilities to enhance the simulation and prediction of physical systems. The second is reservoir computing, a recently popularized approach within the recurrent neural network framework that is characterized by its efficiency in processing complex time-series data.}

\subsubsection{Reservoir Computing}
\add{Reservoir computing represents a significant leap forward in machine learning, combining ideas from recurrent neural networks and echo state networks. At its core, it introduces an efficient way to handle data that changes over time by focusing on a "reservoir" - a complex, high-dimensional structure. Unlike traditional neural networks, which require intensive training of all layers, reservoir computing simplifies the process by training only the output layer, making it much more efficient. This simplicity is particularly beneficial in several fields, such as optoelectronics, as highlighted by Paquot et al. in 2011 \cite{Paquot2011Optoelectronic}.}

\add{In the field of metrology, reservoir computing has led to remarkable improvements, especially in dealing with complex, fluctuating signals where standard techniques struggle. Przyczyna et al. \cite{Przyczyna2020Reservoir} demonstrated its promising role in improving the sensitivity and speed of chemical sensors and measurement systems.}

\add{As an example of its application in metrology, research by Xia et al. \cite{Xia2023Time-Varying} explores quantum reservoir engineering to better estimate the interaction strength in atom-cavity systems. Quantum reservoir computing (QRC) combines the principles of quantum and reservoir computing, using the dynamics of quantum systems to process time-sensitive data. This quantum version takes advantage of quantum properties such as superposition and entanglement, potentially offering superior performance over classical approaches.}

\begin{figure}[bth]
    \centering
    \includegraphics[width=0.75\linewidth]{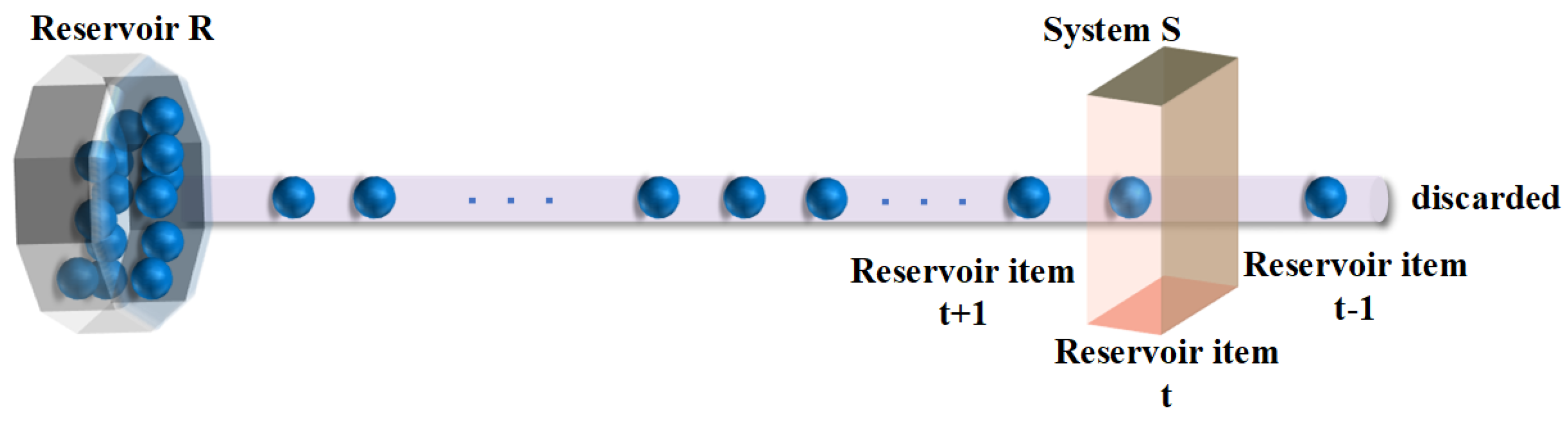}
    \caption{ Framework of the quantum reservoir consisting of a sequence of qubits initialized at the
same state \cite{Xia2023Time-Varying} .}
    \label{fig:enter-label}
\end{figure}

\add{Significant advances in QRC include Nakajima et al.'s strategy of using spatial multiplexing \cite{Nakajima2018Boosting}. This approach uses multiple quantum systems simultaneously, increasing computational power without the need for a large, singular quantum system. This method fits well with current experimental setups, such as those using nuclear magnetic resonance, allowing for more qubits while keeping experiments manageable. In addition, Chen, Nurdin, and Yamamoto in 2020 introduced a quantum reservoir computing model that exploits complex quantum dynamics \cite{Chen2020Temporal}. Their work suggests that even small, imperfect quantum systems, like those in today's quantum computers, can effectively tackle complex, time-based tasks. This discovery opens up the possibility of using emerging quantum computers in a wide range of applications, from neural modeling to language processing.}

\add{By incorporating reservoir computing into metrology, we can overcome the limitations of traditional signal analysis and enable real-time tracking and evaluation of dynamic chemical processes. This advancement not only improves the accuracy and speed of measurement tools, but also paves the way for new diagnostic and sensing technologies that push the boundaries of what is possible in measurement science.
}

\subsubsection{Artificial Neural Twin}

\add{In the field of data-driven metrology, the concept of Artificial Neural Twins (ANT) is a key innovation that bridges the gap between real-time data acquisition and predictive analytics. By mirroring physical systems in a digital framework, ANT enables unprecedented levels of precision and adaptability in measurement technologies. To seamlessly integrate an entire measurement hardware setup into a neural network that can be optimized toward a universal quality criterion, it is essential to ensure that each process step and each sensor is compatible with a state-of-the-art AI interface. This interface is designed with two main goals in mind: first, to facilitate full error backpropagation, allowing deviations from the overall quality benchmark to be tracked and corrected through all stages of the measurement process. The second is to dynamically adjust the settings of each component in the system based on these errors, thereby minimizing deviations from the desired quality standard.}

\add{To achieve these goals, the AI interface must meet several critical criteria. It should be able to compute the partial derivative of each component's output with respect to its input signal, as well as the partial derivative of the component's output with respect to its adjustable parameters. In addition, the interface must preserve the current signal for updating downstream components and compute a local quality metric based on the prevailing material flow. This structured approach allows the precise quantification of each component's contribution to the overall error and enables targeted adjustments to reduce these discrepancies.}

For the practical implementation of the first two properties, the theoretical foundation laid out in \cite{maier2019learning} is utilized. This approach establishes a robust basis for the AI interface, ensuring that each element within the system contributes effectively to achieving the global quality measure.

\add{Building on this foundation, the ANT concept emerges as a key innovation for system adaptation, aiming for superior performance by leveraging advances in AI and digital partnership. The ANT paradigm seeks to transcend conventional methodologies by applying deep learning innovations, particularly through the use of specialized neural network layers, such as convolutional layers, that introduce valuable invariances. These invariances are critical for streamlining the model by reducing the number of parameters and incorporating inherent biases, thereby increasing the efficiency of the network. The main challenge in this area is to identify the most effective arrangement and configuration of these invariances in order to develop powerful deep learning architectures. This task is complex and requires considerable effort. To address this, the concept of known operator learning proposes a blend of analytical models from traditional theories and deep learning approaches, resulting in hybrid or "gray box" models. These models combine data-driven insights with established invariances, effectively bounding errors and reducing the need for large training datasets. This synergy significantly increases the ability of models to generalize, especially in areas where domain expertise is critical, such as signal reconstruction and image processing.}

\begin{figure}[tb]
    \centering
    \includegraphics[width=\linewidth]{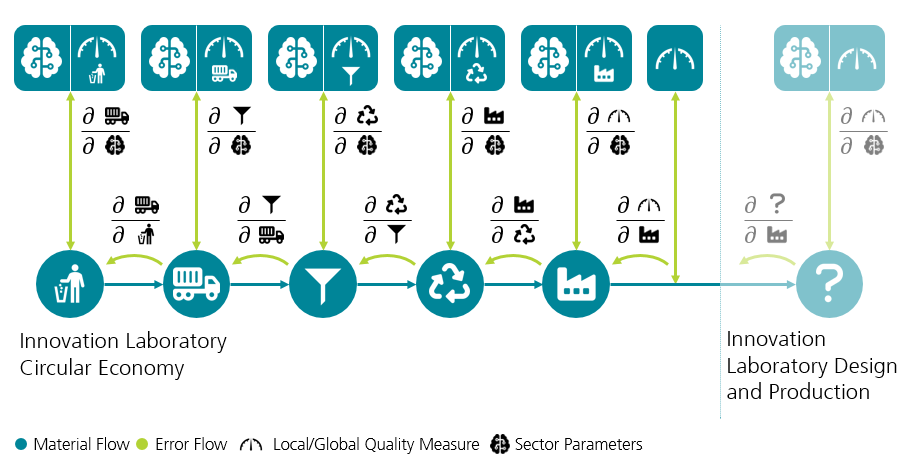}
    \caption{Artificial Neural Twin illustrated in the context of waste management from the K3I-Cycling Project funded by the German Federal Ministry of Education and Research: Each processing step fulfills the requirements for a decentralized AI interface.}
    \label{fig:k3i_cycling}
\end{figure}

The ANT concept is pivotal in the current digital transformation, advocating for the creation of digital twins that accurately reflect physical entities to facilitate their evaluation, optimization, and prediction. The rapid evolution of technologies such as the Internet of Things, Big Data, multi-physics simulation, Industry 4.0, alongside real-time sensors and sensor networks, has catalyzed the growth of digital twins. These virtual models represent a convergence of vast amounts of information from the physical world, necessitating advanced solutions for their processing and synchronization. Machine learning, particularly deep learning, has shown promise in enhancing these digital representations, whether it's through refining data extraction from physical environments or modeling complex systems as black box models when faced with high complexity or scant information. An example of an ANT implementation is given in Figure~\ref{fig:k3i_cycling} for the scenario of a complex waste management system.

The widespread integration of AI within digital twins is still nascent, with its application mainly confined to specific modules or as a tool within limited domains. The vision of a seamless, end-to-end integrated digital twin system is yet to be realized. ANT advocates for a strategic application of black, grey, and white box models to achieve this comprehensive integration. However, one of the barriers to the adoption of ANT is the lack of standardization in digital twinning methodologies. Overcoming this obstacle is essential for the advancement and establishment of ANT as a benchmark for the future of digital and physical world synthesis.

The recent innovation in artificial intelligence, particularly in deep learning, has led to AI methods becoming standard solutions in various domains. However, these methods are often implemented as isolated solutions, addressing single, specific problems, especially in fields like medicine and industrial production. This isolated approach means that AI methods typically do not integrate information from preceding or subsequent steps in a process, such as separating signal reconstruction from AI-based evaluations like classification or segmentation. This limits the potential of AI, as only parts of a process are enhanced, not the entire system. Most AI methods operate as a black box, requiring extensive training data, making solutions heavily data-dependent. To address this, the ANT aims to digitize entire process chains, including all components and material flows. Unlike traditional digital twins, ANT is a fully differentiable digital representation capable of optimizing both individual components and the entire system towards a global quality measure while considering local conditions. A key innovation is the development of a standardized interface for system components, ensuring modularity and compatibility. This interface requires components to provide their own derivations relative to the input, allowing for domain-specific implementation by component manufacturers without revealing proprietary information. This approach ensures the protection of sensitive information and is essential for industry acceptance, overcoming obstacles like the lack of standardization and privacy concerns \cite{Kor2022}, which have previously hindered digitalization and AI innovation in industrial settings.

% Conclusion and outlook

\section{Conclusion and Outlook}

In metrology, the mathematical model that relates the measured values to the calculated value of the measurand plays an essential role. With the help of a model, a relationship can be established between the observed values and the underlying mechanisms. This relationship can be established for simple measurements of a univariate quantity as well as for complex measurements. The theory of mathematical modelling in measurement is well established and a wide range of methods can also be found in the literature. \\
In the future, there will be more and more networked, distributed measurement systems, large amounts of volatile data and data-driven modelling approaches. The combination of these characteristics poses challenges to established measurement modeling approaches. In addition, many sensors are typically combined with other data sources to form a complex network of sensors.\\
In this paper, we have briefly reviewed existing concepts from modeling and discussed their potential use for data analysis scenarios in the future. White-box, black-box and grey-box models are already known in many fields. The use of white- and grey-box models is quite well known in metrology, with e.g. the recent addition to the GUM focusing on these model types. The extension and further development of the underlying principles for handling data-driven black-box models will build on this knowledge. Therefore, this paper sets out the basic principles for white- and grey-box models and shows how they differ from black-box models. Furthermore, this paper offers insight into the versatile applications of data-driven modeling. In particular, adaptive and intelligent autonomous systems, which include the concept of digital twins, are described in detail and their potential use is highlighted using various examples. Due to the complex nature of real sensor networks and the volatility of the measured data, these approaches will combine several of the modeling principles outlined in this paper and will lead to AI systems implemented in distributed hardware as outlined in the ANT concept.

\section{Acknowledgements}
A main part of this work was financed by the \glqq SmartCT - Artificial Intelligence Methods for an Autonomous Robotic CT System\grqq \ project (project nr. DIK-2004-0009). \\
Additionally, part of this work was funded by German Ministry of Education and Research (BMBF) under grant number 033KI201.\\
Also, part of this work has been developed within the Joint Research project 17IND12 Met4FoF of the European Metrology Programme for Innovation and Research (EMPIR). The EMPIR is jointly funded by the EMPIR participating countries within EURAMET and the European Union.

Our special thanks go to our author colleagues with whom we jointly prepared the manuscript \glqq Modelling of Networked Measuring Systems - From White-Box Models to Data Based Approaches\grqq \ \cite{paper} within the EMPIR project 17IND12 Met4FoF, namely Andonovic, I., University of Strathclyde, UK; Dorst, T., ZeMA - Zentrum für Mechatronik und Automatisierungstechnik gGmbH, Saarbrücken; Eichstädt, S., Physikalisch-Technische Bundesanstalt, Braunschweig and Berlin; Füssl, R., Technische Universität Ilmenau; Gourlay, G., University of Strathclyde, UK; Harris, P. M., National Physical Laboratory, Teddington, UK; Heizmann, M., Karlsruhe Institute of Technology; Luo Y., National Physical Laboratory, Teddington, UK; Schütze, A., Saarland University, Saarbrücken;Sommer, K.-D., Ilmenau University of Technology; Tachtatzis, C., University of Strathclyde, UK.

\newpage

% -----------------------------
%\bibliographystyle{alphadin} 
\bibliographystyle{unsrt}
\bibliography{literatur.bib}
\end{document}